\documentclass[lettersize,journal]{IEEEtran}
\usepackage{amsmath,amsfonts}
\usepackage{array}
\usepackage[caption=false]{subfig}
\usepackage{textcomp}
\usepackage{stfloats}
\usepackage{url}
\usepackage{verbatim}
\usepackage{graphicx}
\usepackage{cite}
\usepackage{pifont}
\usepackage[noend]{algpseudocode}

\usepackage{amsmath}
\usepackage{amssymb}
\usepackage{booktabs}
\usepackage{multirow}
\usepackage[pagebackref,breaklinks,colorlinks]{hyperref}

\begin{document}

\title{Large Separable Kernel Attention: Rethinking the Large Kernel Attention Design in CNN}

\author{Kin Wai Lau$^{1,2}$, Lai-Man Po$^{1}$, Yasar Abbas Ur Rehman$^{2}$ \\
City University of Hong Kong$^{1}$\\
TCL AI Lab$^{2}$\\
\thanks{K.W. Lau is with the Department of Electrical Engineering, City University of Hong Kong, Hong Kong, and also with TCL AI Lab. Y.A.U. Rehman is with TCL AI Lab. (e-mail: kinwailau6-c@my.cityu.edu.hk, yasar.abbas@my.cityu.edu.hk)}
\thanks{L.-M. Po is with the Department of Electrical Engineering, City University of Hong Kong, Hong Kong (email: eelmpo@cityu.edu.hk)}}

\markboth{Journal of \LaTeX\ Class Files,~Vol.~14, No.~8, August~2023}%
{Shell \MakeLowercase{\textit{et al.}}: A Sample Article Using IEEEtran.cls for IEEE Journals}


\maketitle
\newcommand{\cmark}{\ding{51}}
\newcommand{\xmark}{\ding{55}}

\begin{abstract}
Visual Attention Networks (VAN) with Large Kernel Attention (LKA) modules have been shown to provide remarkable performance, that surpasses Vision Transformers (ViTs), on a range of vision-based tasks. However, the depth-wise convolutional layer in these LKA modules incurs a quadratic increase in the computational and memory footprints with increasing convolutional kernel size.  To mitigate these problems and to enable the use of extremely large convolutional kernels in the attention modules of VAN, we propose a family of Large Separable Kernel Attention modules, termed LSKA. LSKA decomposes the 2D convolutional kernel of the depth-wise convolutional layer into cascaded horizontal and vertical 1-D kernels. In contrast to the standard LKA design, the proposed decomposition enables the direct use of the depth-wise convolutional layer with large kernels in the attention module, without requiring any extra blocks. We demonstrate that the proposed LSKA module in VAN can achieve comparable performance with the standard LKA module and incur lower computational complexity and memory footprints. We also find that the proposed LSKA design biases the VAN more toward the shape of the object than the texture with increasing kernel size. Additionally, we benchmark the robustness of the LKA and LSKA in VAN, ViTs, and the recent ConvNeXt on the five corrupted versions of the ImageNet dataset that are largely unexplored in the previous works. Our extensive experimental results show that the proposed LSKA module in VAN provides a significant reduction in computational complexity and memory footprints with increasing kernel size while outperforming ViTs, ConvNeXt, and providing similar performance compared to the LKA module in VAN on object recognition, object detection, semantic segmentation, and robustness tests. Codes are available at \url{https://github.com/StevenLauHKHK/Large-Separable-Kernel-Attention}.
\end{abstract}


\section{Introduction}
\label{sec:intro}
Over the past decade, Convolutional Neural Network (CNN) architectures and optimization techniques have expeditiously evolved. This evolution comes from the designing of activation functions \cite{xu2015empirical, agarap2018deep, jang2016categorical}, proposing regularization methods of CNN parameters \cite{srivastava2014dropout, santurkar2018does, cortes2012l2}, constructing new optimization methods \cite{kingma2014adam, loshchilov2017decoupled, foret2020sharpness}, cost functions \cite{hadsell2006dimensionality, schroff2015facenet, wang2014learning}, and new network architectures \cite{krizhevsky2017imagenet, szegedy2016rethinking, he2016deep, chollet2017xception, xie2017aggregated}. Most of these breakthroughs in CNN are clustered around the human cognitive processes, specifically the human visual system.

Owing to translational equivariance  and locality  properties, CNN is a common choice of a feature encoder for various vision-based tasks, including image classification \cite{krizhevsky2017imagenet, szegedy2016rethinking, he2016deep}, semantic segmentation \cite{takikawa2019gated, yu2018bisenet, mostajabi2015feedforward, zhou2018unet++}, and object detection \cite{ren2015faster, he2015spatial, dai2016r, liu2016ssd}. Further improvement in the performance of CNN, for these vision-based tasks, is obtained by using attention mechanisms.
\begin{figure}[t]
\centering
\includegraphics[width=\linewidth]{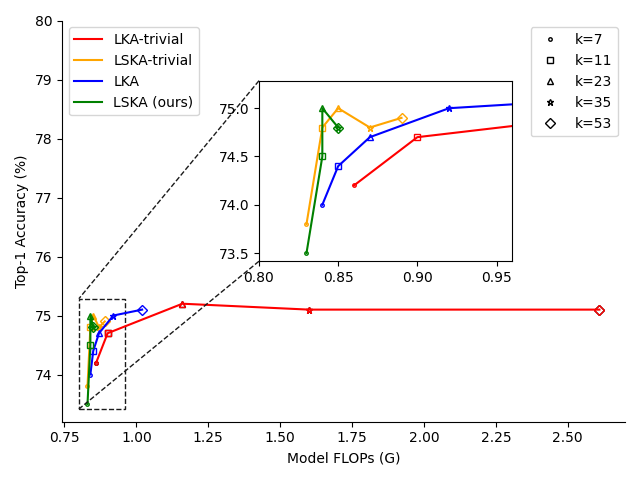}
\caption{Speed-accuracy trade-off under different large kernel decomposition methods and kernel sizes. Different markers represent different kernel sizes and VAN-Tiny is used in our comparison. We can see that the naive design of LKA (LKA-trivial) and the actual design of LKA in VAN result in higher GFLOPs with increasing kernel size. In comparison, our proposed LSKA-trivial and LSKA in VAN significantly reduce the GFLOPs without downgrading the performance. }
\label{fig:flops-accuracy-classification}
\end{figure}
For instance, the authors in \cite{woo2018cbam, hu2018squeeze, wang2018non, jaderberg2015spatial, mnih2014recurrent} have shown that the attention mechanisms improve the performance of CNN by producing salient regions that are consistent with human eyes and brain. Although CNN with attention mechanisms are being effective, the recent surge in the adaptation of self-attention-based networks in the vision domain, such as Vision Transformer (ViT) \cite{dosovitskiy2020vit} and its variants \cite{liu2021swin} has outperformed the CNN on image classification, object detection, and semantic segmentation. The superior performance of ViTs compared to CNN is attributed to the better scaling behavior of Transformers that uses Multi-Head-Self-Attention (MHSA) as its key component \cite{liu2022convnet}. However, the effectiveness of ViTs in the vision domain comes at the cost of a quadratic increase in the computational and memory footprints when presented with high-resolution input images. Nevertheless, the effectiveness of ViTs such as Swin-Transformer \cite{liu2021swin} has reopened the doors for the re-exploration of the design choices of CNN that were under the shadow since the introduction of VGG network \cite{liu2022convnet}.

One of the main reasons behind the success of ViTs compared to CNN in image classification is its ability to model long-range dependency of the input image. Such long-range dependency in CNN can be modeled by using larger receptive fields with  attention mechanisms \cite{guo2022visual}. The large receptive fields in CNN are obtained via either stacking many convolutional operations \cite{wang2018non} or using a larger kernel size \cite{ding2022scaling}. While the former approach may lead to an increase in the size of the model, the latter approach is thought to be cost-prohibitive due to its memory and computational requirements.
However, in a recent study, the authors show that the large kernel in CNN can be modeled by connecting in cascade depth-wise convolution and dilated depth-wise convolution without incurring a quadratic rise in the computational and memory footprints \cite{guo2022visual}. Their proposed Visual Attention Network (VAN) uses a stack of kernels with a simple attention mechanism, named Large Kernel Attention (LKA) as shown in Fig. \ref{fig:LKA-vs-LSKA-c}. The LKA module adopts a standard depth-wise convolution with small receptive field kernels to capture the local dependencies and compensate for the gridding issues \cite{wang2018understanding} followed by a dilated depth-wise convolution with a large receptive field kernel to model long-range dependencies. Such a combination of depth-wise convolution and dilated depth-wise convolution is equivalent to the large-scale CNN kernels as proposed in \cite{liu2022convnet,ding2022scaling}. The output of the dilated depth-wise convolution is fed to the $1\times1$ convolution to infer the attention map. The output attention maps are then multiplied by the input feature for adaptive feature refinement.

The VAN with LKA has been shown to outperform the state-of-the-art ViTs and CNNs in image classification, object detection, and semantic segmentation. However, as pointed out in \cite{ding2022scaling}, the naive design of large-scale depth-wise convolutional kernel still incurs high computational and memory footprints downgrading the effectiveness of the model with increasing kernel size. Our preliminary results, as shown in Fig. \ref{fig:flops-accuracy-classification}, are in inline with the preliminary results of \cite{ding2022scaling}, where we found that the design depth-wise convolution in the LKA module (without using depth-wise dilated convolution) of VAN is computationally inefficient with the kernel sizes as large as $35\times35$ and $53\times53$. 

In this paper,  we first study the effect of depth-wise convolution with a simple attention module in VAN against large kernel sizes. We termed such design of depth-wise convolution with large kernel and the attention module as LKA-trivial. Secondly, we proposed a separable version of the depth-wise convolution for VAN. The separable version of depth-wise convolution evenly divides a given $k\times k$ convolutional kernel into $1\times k$  and $k\times 1$ separable convolutional kernels that act in a cascade on the input features. Keeping other factors constant, the proposed separable version of the depth-wise convolution in the LKA-trivial module of VAN significantly reduces the quadratic growth in the number of parameters with increasing kernel size. We termed this modified design of LKA-trivial as LSKA-trivial. Moreover, this type of kernel decomposition is also compatible with the depth-wise dilated convolution allowing us to propose a completely separable version of the LKA module \cite{guo2022visual} in VAN. We termed this proposed separable version of the LKA module as LSKA. We show that the proposed LSKA version of LKA obtains similar performance in VAN while being computationally efficient even at larger kernel sizes. Additionally, the proposed LSKA module at larger kernel sizes enhances the long-range dependency of the input image without incurring high computational and memory footprints. 

To provide a formal illustration of the effectiveness of the proposed LSKA in VAN, we extensively evaluate it against LKA in VAN on a range of downstream tasks. We also investigate the robustness of the proposed LSKA and other baselines, like LKA in VAN, ViTs \cite{liu2021swin, wang2022pvt, wang2021pyramid, touvron2021training}, and ConvNeXt \cite{liu2022convnet}, on various distortion datasets like common corruptions, semantic shift, and out-of-distribution natural adversarial examples that remain unexplored in previous works. The contribution of our work can be summarized as follows:

\begin{enumerate}
    \item We address the issue of computational inefficiency of the depth-wise convolutional kernel with increasing kernel size for LKA-trivial and LKA in VAN. We show that replacing the $k\times k$ convolutional kernel in depth-wise convolution with a cascaded $1\times k$ and $k\times 1$ convolutional kernel effectively reduces the quadratic growth in the number of parameters incurred by LKA-trivial and LKA in VAN with increasing kernel size, without any performance degradation. 
    
    \item We experimentally validate the effectiveness of LSKA in the VAN on various  vision-based tasks that include image classification, object detection, and semantic segmentation. We demonstrate that LSKA can benefit from a large kernel while maintaining the same inference-time costs compared to a small kernel in the original LKA.
    
    \item We benchmark the robustness of LKA-based VAN, LSKA-based VAN, ConvNeXt, and state-of-the-art ViTs on 5 diverse ImageNet datasets, which contain various types of perturbations applied to the images. Our results indicate that LSKA-based VAN is a robust learner compared to previous large kernel CNNs and ViTs.   
    \item We provide quantitative evidence to show that the features learned by the large kernel in LSKA-based VAN encode more shape information and less texture compared to ViTs and previous large kernel CNNs. In addition, there is a high correlation between the quantity of shape information encoded in the feature representations and robustness against different image distortions. This evidence helps us explain why LSKA-based VAN is a robust learner.
    
\end{enumerate}
The rest of this paper is organized as follows. In Section \ref{sec:related-work} we introduce the state-of-the-art work done on large kernel design and attention networks. Section \ref{sec:method} presents our proposed LSKA design for VAN followed by the experiment results in Section \ref{sec:experiment}. Section \ref{sec:ablation-studies} provides ablation studies with different large kernel decomposition methods and kernel sizes. Section \ref{sec:sota} makes a comparison with the state-of-the-art model. Section \ref{sec:sota-robust} shows the robustness comparison between LSKA-VAN, CNNs, and ViTs.  Finally, we conclude our work in Section \ref{sec:conclusion}.

\section{Related Work}
\label{sec:related-work}
\subsection{CNNs with Large Kernels}
\label{sec:cnn_large_kernel}
A plethora of research work was done in the last decade to improve the Convolutional Neural Network (CNN) architectures for general image recognition tasks. Except for AlexNet \cite{krizhevsky2017imagenet} and Inception \cite{szegedy2016rethinking} networks, the kernel design in these CNN architectures was predominantly limited to $3\times3$ (e.g., VGG \cite{DBLP:journals/corr/SimonyanZ14a}, ResNet \cite{he2016deep}, MobileNets \cite{DBLP:journals/corr/HowardZCKWWAA17}), owing to its computational efficiency with increasing depth of the weight layers. In an attempt to use a large kernel size, the  authors in \cite{hu2019local} proposed a Local Relation Network (LR-Net) made up of local relational layers with a relatively large kernel size of 7$\times$7. Although achieving better performance than the conventional ResNets (with the same kernel size), the performance of LR-Net dropped when the kernel size was increased further. To bridge the performance gap between hierarchical transformers and CNN, ConvNeXt \cite{liu2022convnet} conducts an empirical study by gradually borrowing the transformer designs into the ResNet. They discover several key components that improve the CNN performance such as changing the training procedure as in Swin Transformer, changing the stage compute ratio, using fewer activation and normalization layers, and using a larger kernel size. Similar to LR-Net, they showed that the performance saturates when the kernel size is increased beyond $7\times7$. 

Recently, the authors in \cite{ding2022scaling} revisited the large kernel design in CNNs that was the under the shadow for a long time. They demonstrated that replacing the set of small-weight kernels with a few reparametrized large-weight kernels in MobileNetV2 resulted in large Effective Receptive Fields (ERFs) and partially mimic the human-like understanding of the shape of the object. Their proposed RepLKNet, with a large kernel size of $31\times31$, outperformed Swin Transformers by $0.3\%$ on ImageNet classification and ResNet-101 by $4.4\%$ on MS-COCO detection. However, RepLKNet incurs high computational footprints thus limiting its effectiveness in other domains such as segmentation. For example, the authors in \cite{peng2017large} show that the naïve large kernel convolution harms the performance of segmentation tasks because the model suffers from an overfitting problem as the parameter size increases with the kernel size. To overcome this issue, they proposed 
Global Convolution Network (GCN) \cite{peng2017large} with a large 1$\times k$ and $k\times$1 convolutional kernel to improve semantic segmentation performance.

Another recent work, SLaK \cite{liu2023more}, observe that the performance of RepLKNet \cite{ding2022scaling} begins to plateau as the kernel size increases beyond 31, 51, and 61. To address the trainability of an extremely large kernel, SLaK  decomposes the large kernel into two rectangular kernels (i.e., $51 \times 5$ and $5 \times 51$) and uses the dynamic sparsity technique to reduce the learnable parameters. Unlike these methods, we employ separable kernels in depthwise convolution and depthwise-dilated convolution with an attention module for the CNN-based Visual Attention Network (VAN) \cite{guo2022visual} to further improve its computational efficiency.

\subsection{Attention Mechanisms with Large Kernel}
\label{sec:Attention}
Attention Mechanisms are used to select the most important regions in an image. Generally, they can be grouped into four categories: spatial attention \cite{wang2018non,jaderberg2015spatial, dosovitskiy2020vit, woo2018cbam}, channel attention \cite{hu2018squeeze,wang2020eca, woo2018cbam}, temporal attention \cite{xu2017jointly, zhang2019scan}, and branch attention \cite{li2019selective,zhang2022resnest}. Here, we focus more on channel attention and spatial attention as they are more related to our work. Channel attention focuses on the "what" semantic attribute of the particular model's layer. As each channel of the feature map act as a response map of a detector, also known as filter \cite{chen2017sca}, the channel attention mechanism allows the model to focus on specific attributes of the object across the channels \cite{hu2018squeeze}. 
In contrast to channel attention, spatial attention focuses on "where" semantically-related regions the model should pay attention to. STN \cite{jaderberg2015spatial}, GENet \cite{hu2018gather}, and Non-local Neural Networks \cite{wang2018non} are some of the representative works that involve different kinds of spatial attention methodologies.   
\begin{figure*}[t]
      \centering
      \subfloat[LKA-trivial]{\includegraphics[width=0.15\linewidth, height=0.25\linewidth]{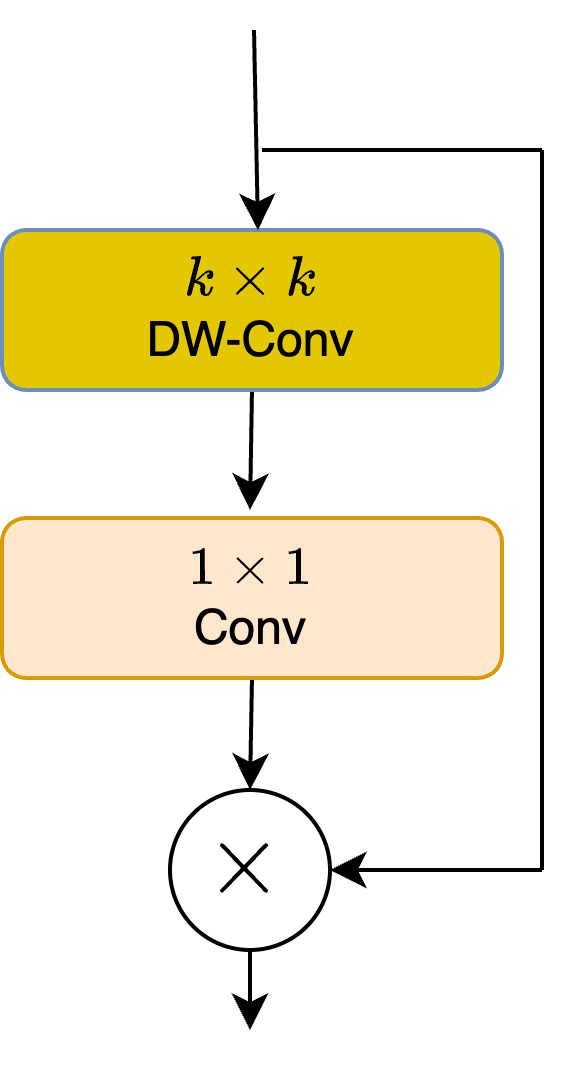} 
      \label{fig:LKA-vs-LSKA-a}} \hspace{0.5cm}
       \subfloat[LSKA-trivial]{\includegraphics[width=0.15\linewidth, height=0.3\linewidth]{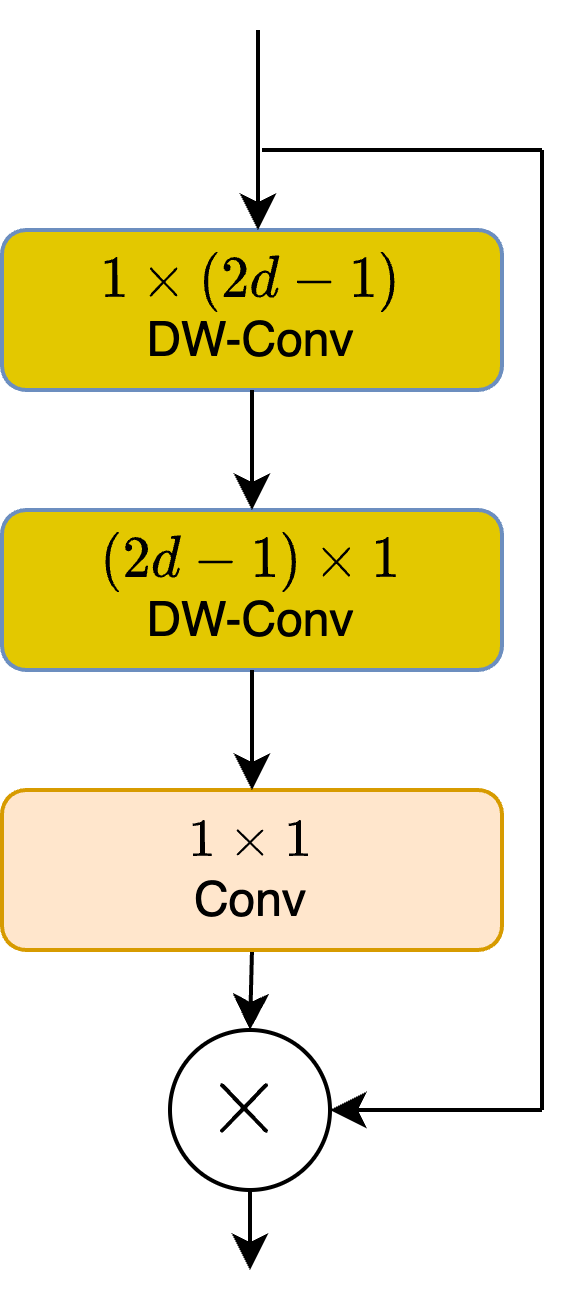}
        \label{fig:LKA-vs-LSKA-b}} \hspace{0.5cm}
        \subfloat[LKA]{\includegraphics[width=0.15\linewidth, height=0.3\linewidth]{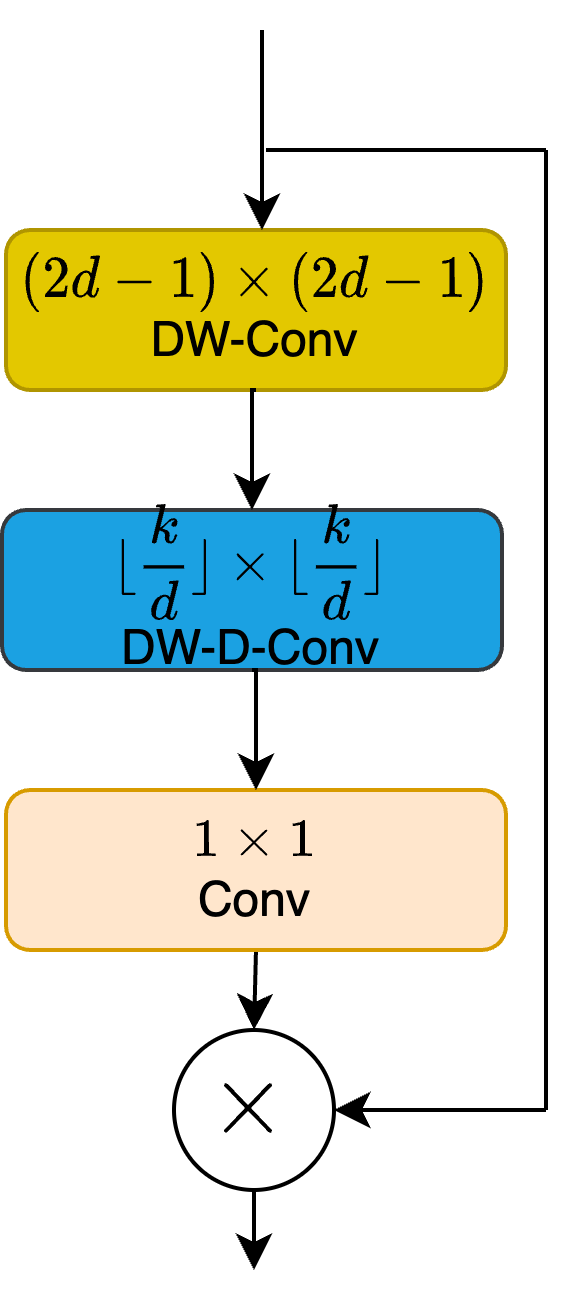}
        \label{fig:LKA-vs-LSKA-c}} \hspace{0.5cm}
        \subfloat[LSKA]{\includegraphics[width=0.15\linewidth, height=0.4\linewidth]{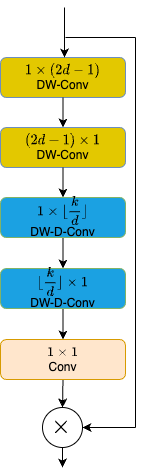}
          \label{fig:LKA-vs-LSKA-d}} 
      \caption{Comparsion on different designs of large kernel attention module. (a) Trivial 2D large kernel depth-wise convolution (DW-Conv) with $1\times1$ convolution (b) Cascaded horizontal and vertical 1D large kernel depth-wise convolution with $1\times1$ convolution (c) Original LKA design in VAN including a standard depth-wise convolution (DW-Conv), a dilated depth-wise convolution (DW-D-Conv), and a $1\times1$ convolution. (d) Our proposed LSKA decomposed the first two layers of LKA into four layers, and each layer of LKA is formed by two 1D convolution layers. Notice that $\bigotimes$ represents Hadamard product, k represents the maximum receptive field, and d represents the dilation rate.}
      \label{fig:LKA-vs-LSKA}
\end{figure*}
\begin{figure*}[t]
\centering
      \subfloat[]{\includegraphics[width=0.5\linewidth]{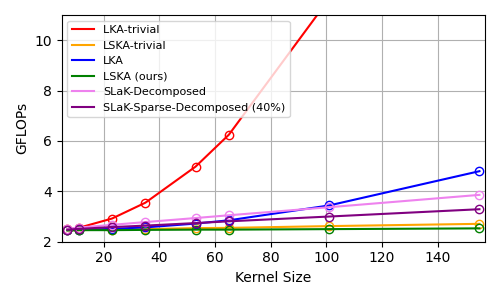}
      \label{fig:acc-vs-kernel-ablation}}
      \subfloat[]{\includegraphics[width=0.5\linewidth]{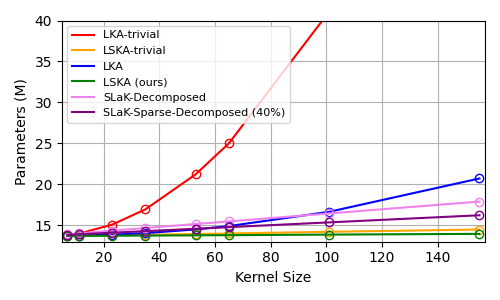}
        \label{fig:param-vs-kernel-ablation}}\\
\caption{Memory and computational complexity comparison between our proposed method, LKA and SLaK methods. (a) The number of GFLOPs with the increasing scale of kernel size. (b) The number of network parameters with the increasing scale of the kernel size. Note that all benchmarks are based on VAN-Small network architecture.} 
\label{fig:MemComplex}
\end{figure*}
Self-Attention \cite{vaswani2017attention}, a type of spatial attention, has been used in CNNs and ViTs \cite{dosovitskiy2020vit, liu2021swin}. 
Visual Attention Network (VAN) \cite{guo2022visual} proposes a new CNN backbone network, which combines CNN's properties and self-attention modules. The author adopts the CNN network architecture with a Large Kernel Attention (LKA) module to achieve the locality of CNN, long-range dependence, and spatial adaptability properties of self-attention modules, similar to ViTs. Additionally, the LKA modules have channel adaptability that is not naturally present in both standard CNN and self-attention modules in transformers. To improve computational efficiency, LKA adopted dilated convolution \cite{DBLP:journals/corr/YuK15} with depthwise convolution \cite{DBLP:journals/corr/HowardZCKWWAA17} (DW-D-Conv) to achieve a larger ERFs.

Even VAN achieves a better performance than a series of transformer networks such as PVT-V2 \cite{wang2022pvt}, Swin Transformer \cite{liu2021swin}, and Twins-SVT \cite{chu2021twins} in image classification, object detection, and semantic segmentation, our work demonstrates that LSKA can further reduce the computational complexity of VANs without any loss of performance.

\section{Methodology}
\label{sec:method}
In this Section, we first discuss how the LKA module (with and without using dilated depth-wise convolution) can be restructured by using 1D convolutional kernels to design the LSKA module. We then summarize several key properties of the LSKA module followed by the complexity analysis of LSKA.
\subsection{Formulation}

We begin by designing the basic LKA block, without using dilated depth-wise convolution, as shown in Fig. \ref{fig:LKA-vs-LSKA-a}. Given an input feature map $F\in\mathbb{R}^{C\times H\times W}$,  where $C$ is the number of input channels, $H$ and $W$ represents the height and width of the feature map respectively, the trivial way to design the LKA is using a large convolutional kernel in 2D depth-wise convolution. The output of the LKA can be obtained by using Eq. \ref{eq:trivial_LKA}-\ref{eq:trivial-LKA-O}.
\begin{equation}
Z^{C} =\sum_{H,W}{W}^{C}_{k \times k} ~* ~F^{C},
    \label{eq:trivial_LKA}
\end{equation}
\begin{equation}
    A^{C}= W_{1\times1} * Z^{C}
\end{equation}
\begin{equation}
    \Bar{F}^{C}=A^{C}\otimes F^{C}.
    \label{eq:trivial-LKA-O}
\end{equation}
Where $*$ and $ \otimes $ represent convolution and Hadamard product, respectively. $Z^{C}$ is the output of the depth-wise convolution obtained by convolving the kernel $W$ of size $k\times k$ with the input feature map $F$. It should be noted that each channel $C$ in $F$ is convolved with its corresponding channel in kernel $W$ following \cite{DBLP:journals/corr/HowardZCKWWAA17}. The $k$ in Eq. \ref{eq:trivial_LKA} also represents the maximum receptive field of the kernel $W$. The output of the depth-wise convolution is then convolved $1\times1$ kernel to obtain the attention map $A^{C}$. The output $\Bar{F}^{C}$ of the LKA is the Hadamard product of the attention map$A^{C}$ and input feature map $F^{C}$. One can see that the depth-wise convolution in the LKA module will incur a quadratic increase in the computational cost with increasing kernel size. We called this design LKA-trivial in this work to differentiate it from the actual design of LKA as mentioned in \cite{guo2022visual}. One can quickly find that increasing the kernel size in LKA-trivial will incur a quadratic ($k^{2}$) increase in the computational complexity in VAN (see Fig. \ref{fig:MemComplex}).

To mitigate the issues of high computation cost of depth-wise convolution at larger kernel sizes in the LKA-trivial, the author in \cite{guo2022visual} decomposes the depth-wise convolution with a large kernel into depth-wise convolution with a small kernel followed by dilated depth-wise convolution with a fairly large kernel (Fig. \ref{fig:LKA-vs-LSKA-c}). Such type of large kernel decomposition helps to mitigate the issue of quadratic increase in the computation cost incurred by the depth-wise convolution alone with the large kernel size. As mentioned in \cite{guo2022visual}, the output of LKA can be obtained as follows.  

\begin{equation}
\Bar{Z}^{C} =\sum_{H,W}{W}^{C}_{(2d-1) \times (2d-1)} ~* ~F^{C}
    \label{eq:LSC}
\end{equation}
\begin{equation}
Z^{C} =\sum_{H,W}{W}^{C}_{\lfloor\frac{k}{d}\rfloor \times \lfloor\frac{k}{d}\rfloor} ~* ~\Bar{Z}^{C}
    \label{eq:LKSC}
\end{equation}
\begin{equation}
     A^{C}= W_{1\times1} * Z^{C},
\end{equation}
\begin{equation}
    \Bar{F}^{C}=A^{C}\otimes F^{C}.
\end{equation}
Where $d$ is the dilation rate. 
$\Bar{Z}^{C}$ in Eq. \ref{eq:LSC} denotes the output of depthwise-convolution with kernel size $\left(2d-1\right) \times \left(2d-1\right)$ which captures the local spatial information and compensates gridding effect \cite{wang2018understanding} of the following depth-wise dilated convolution (Eq. \ref{eq:LKSC}) with kernel size $(\lfloor\frac{k}{d}\rfloor \times \lfloor\frac{k}{d}\rfloor)$. Note that $\lfloor.\rfloor$ represents the floor operation. The dilated depth-wise convolution is responsible for capturing the global spatial information of the output $\Bar{Z}^{C}$ of depth-wise convolution. Although the LKA design improves the LKA-trivial by a large margin, it still incurs high computational complexity and memory footprints in VAN when the kernel size is increased beyond $23\times23$ (see Fig. \ref{fig:MemComplex}). 

\subsection{Large Separable Kernel with Attention}
An equivalent and improved configuration of LKA can be obtained by splitting the 2D weight kernels of depth-wise convolution and depth-wise dilated convolution into two cascaded 1D separable weight kernels. We call this modified configuration of the LKA module as LSKA, as shown in Fig. \ref{fig:LKA-vs-LSKA-d}. Following \cite{guo2022visual}, the output of LSKA can be obtained as follows.

\begin{equation}
\Bar{Z}^{C} =\sum_{H,W}{W}^{C}_{(2d-1) \times 1} * (\sum_{H,W}{W}^{C}_{1\times(2d-1)}  ~* ~F^{C}) 
\end{equation}
\begin{equation}
Z^{C} =\sum_{H,W}{W}^{C}_{\lfloor\frac{k}{d}\rfloor \times 1} * (\sum_{H,W}{W}^{C}_{1\times \lfloor \frac{k}{d} \rfloor}  ~* ~\Bar{Z}^{C}) 
\end{equation}
\begin{equation}
     A^{C}= W_{1\times1} * Z^{C},
\end{equation}
\begin{equation}
     \Bar{F}^{C}=A^{C}\otimes F^{C}.
\end{equation}

The separable version of the LKA-trivial can be obtained in a similar fashion, which we named LSKA-trivial as shown in Fig. \ref{fig:LKA-vs-LSKA-b}. One can see from Fig. \ref{fig:MemComplex}, that both LSKA-trivial and LSKA significantly reduce the computational complexity of VAN compared to  LKA-trivial and LKA. In the following subsections, we report the properties of LSKA that differentiate it from the generic convolution, self-attention, and LKA.       

\begin{figure*}[ht]
      \centering
      \subfloat[VAN-LSKA-Tiny (k=7)]{\includegraphics[width=0.25\linewidth, height=0.25\linewidth]{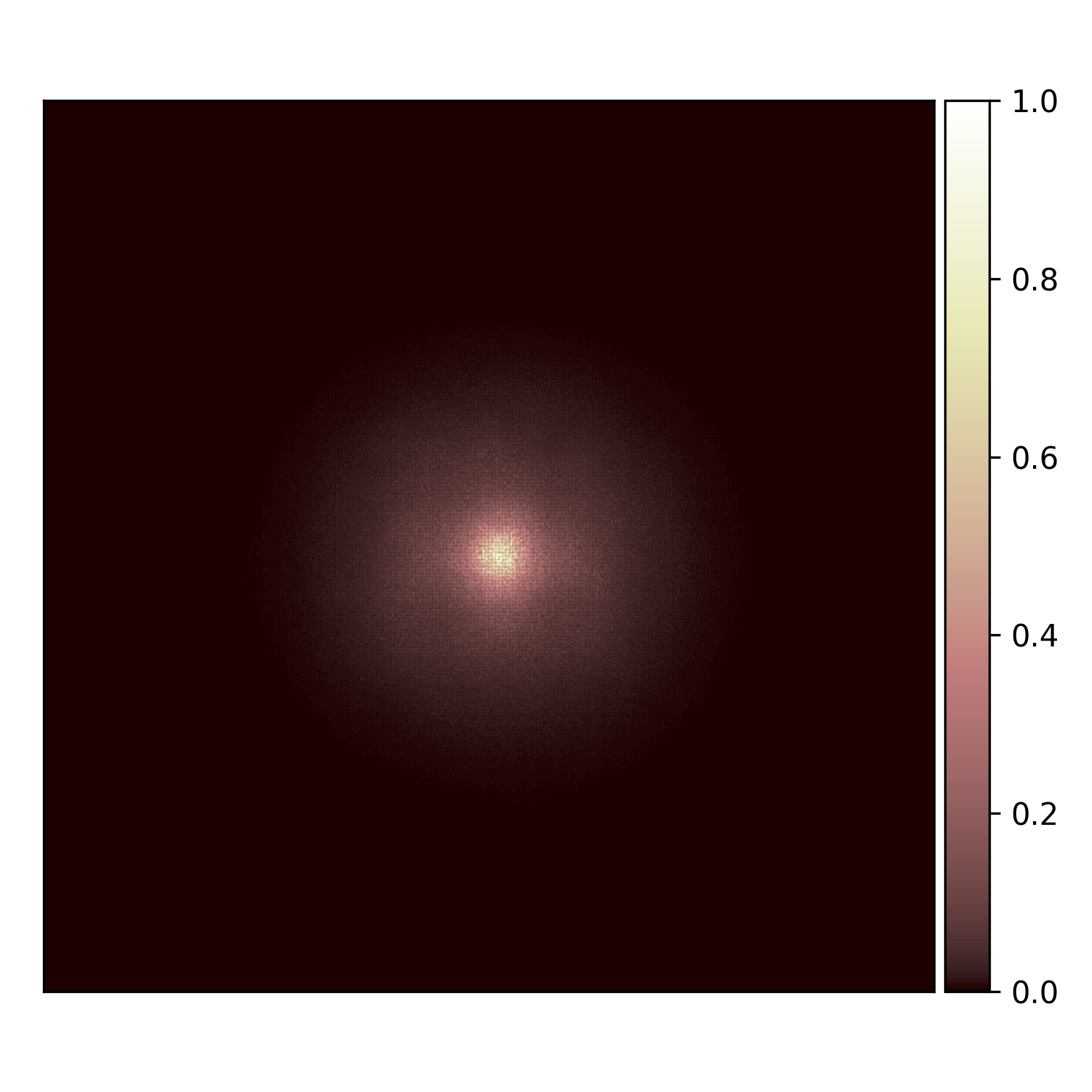}
      \label{fig:erf_k7}}
      \subfloat[VAN-LSKA-Tiny (k=11)]{\includegraphics[width=0.25\linewidth,height=0.25\linewidth]{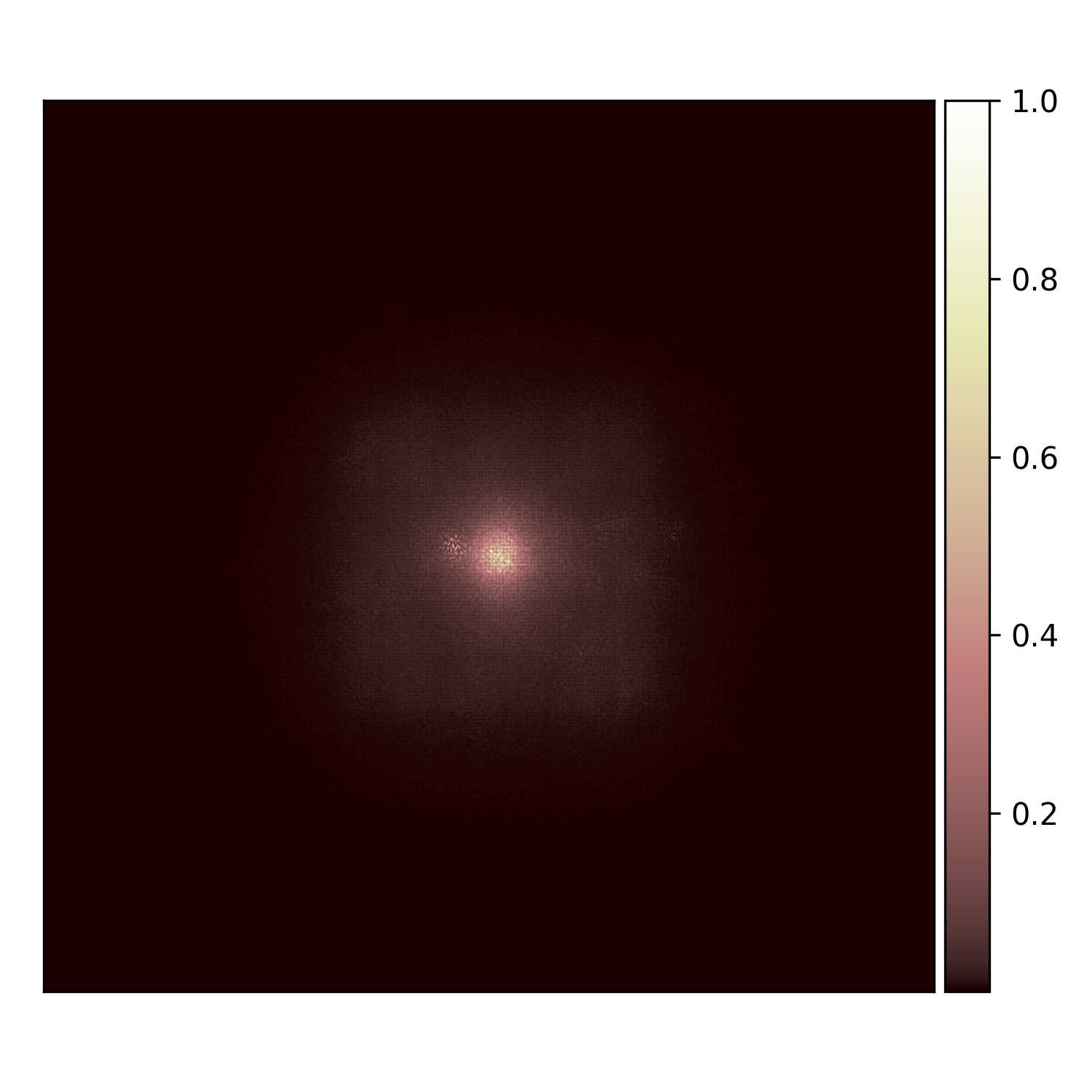}
        \label{fig:erf_k11}}
      \subfloat[VAN-LSKA-Tiny (k=23)]{\includegraphics[width=0.25\linewidth,height=0.25\linewidth]{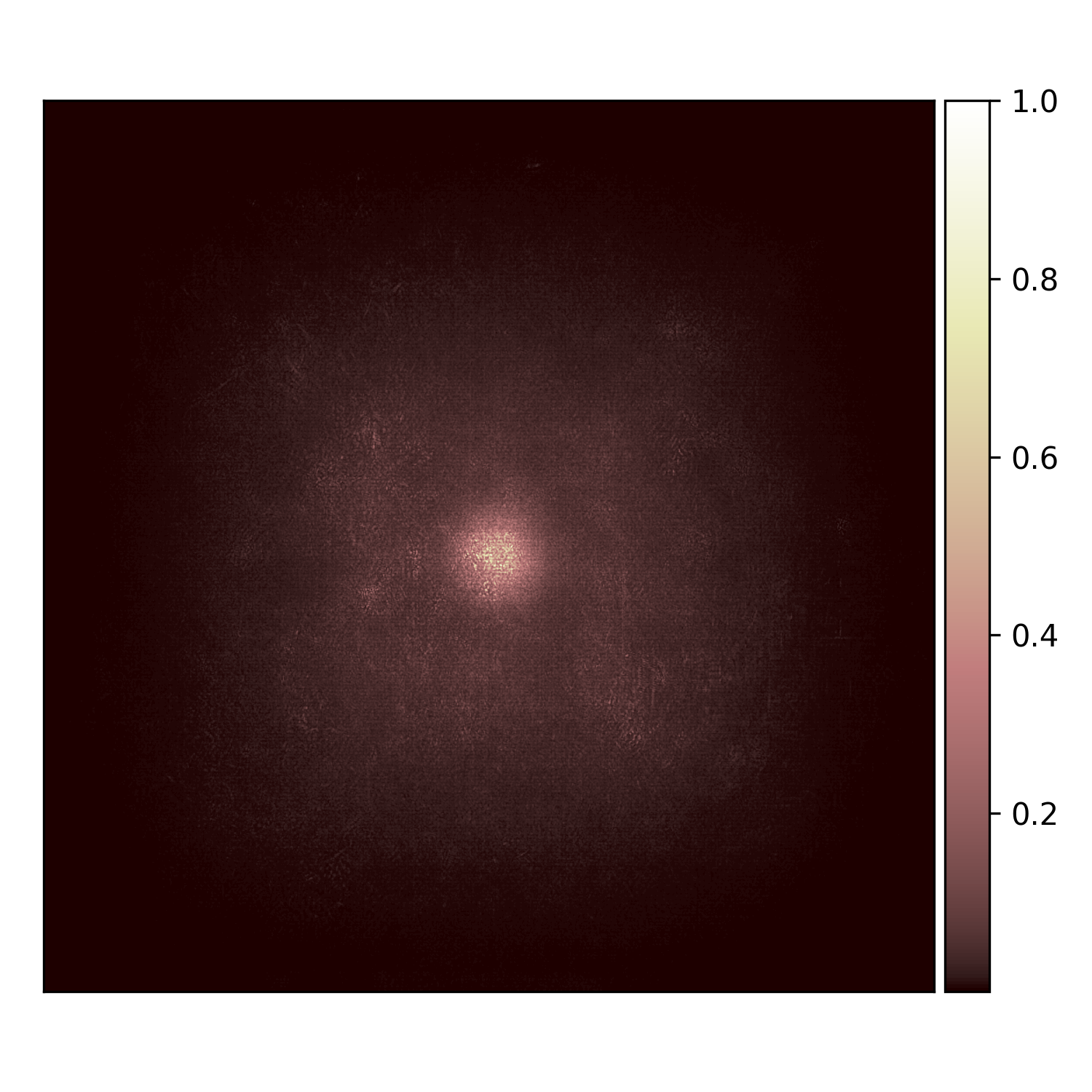}
        \label{fig:erf_k23}}\\
      \subfloat[VAN-LSKA-Tiny (k=35)]{\includegraphics[width=0.25\linewidth,height=0.25\linewidth]{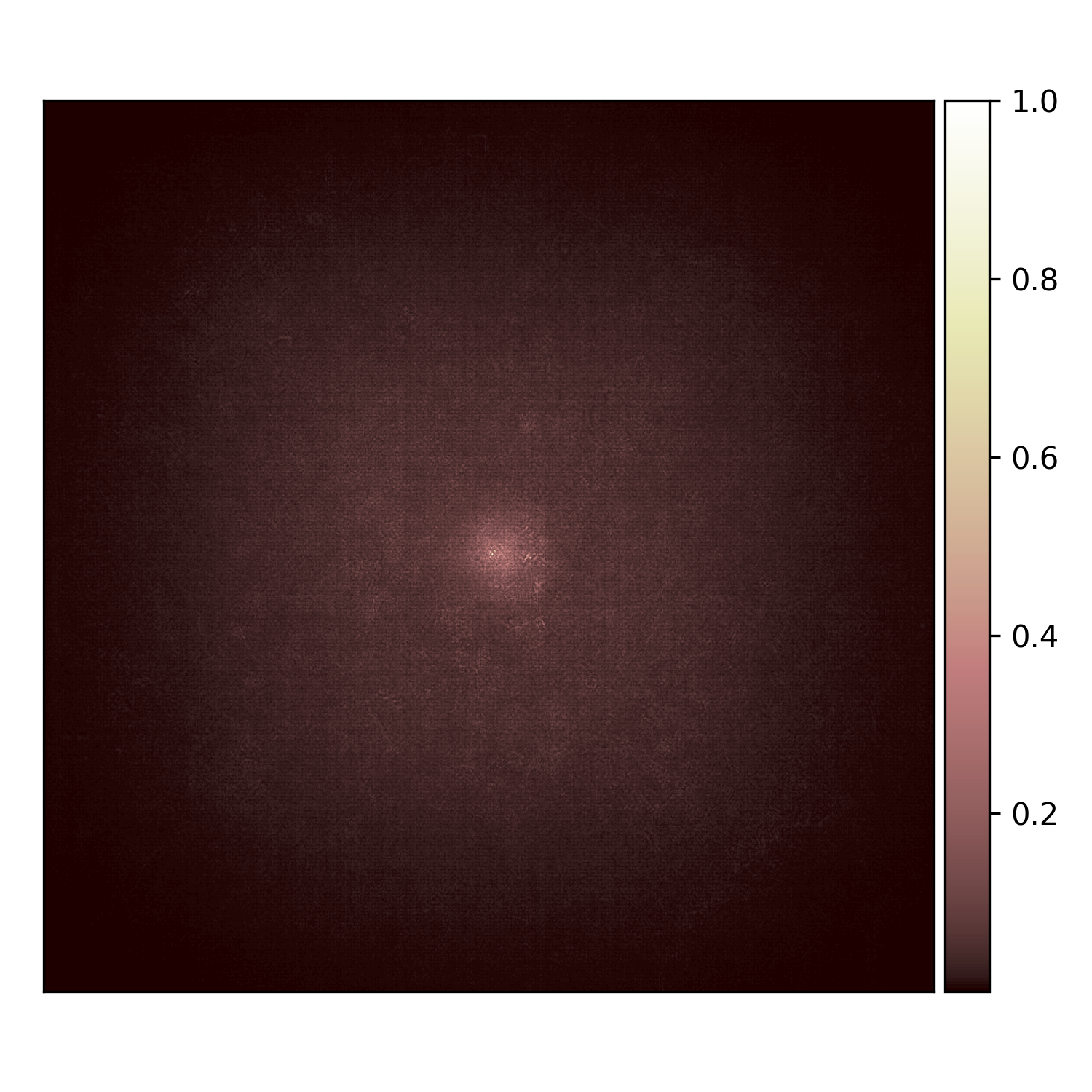}
          \label{fig:erf_k35}
        }
        \subfloat[VAN-LSKA-Tiny (k=53)]{\includegraphics[width=0.25\linewidth,height=0.25\linewidth]{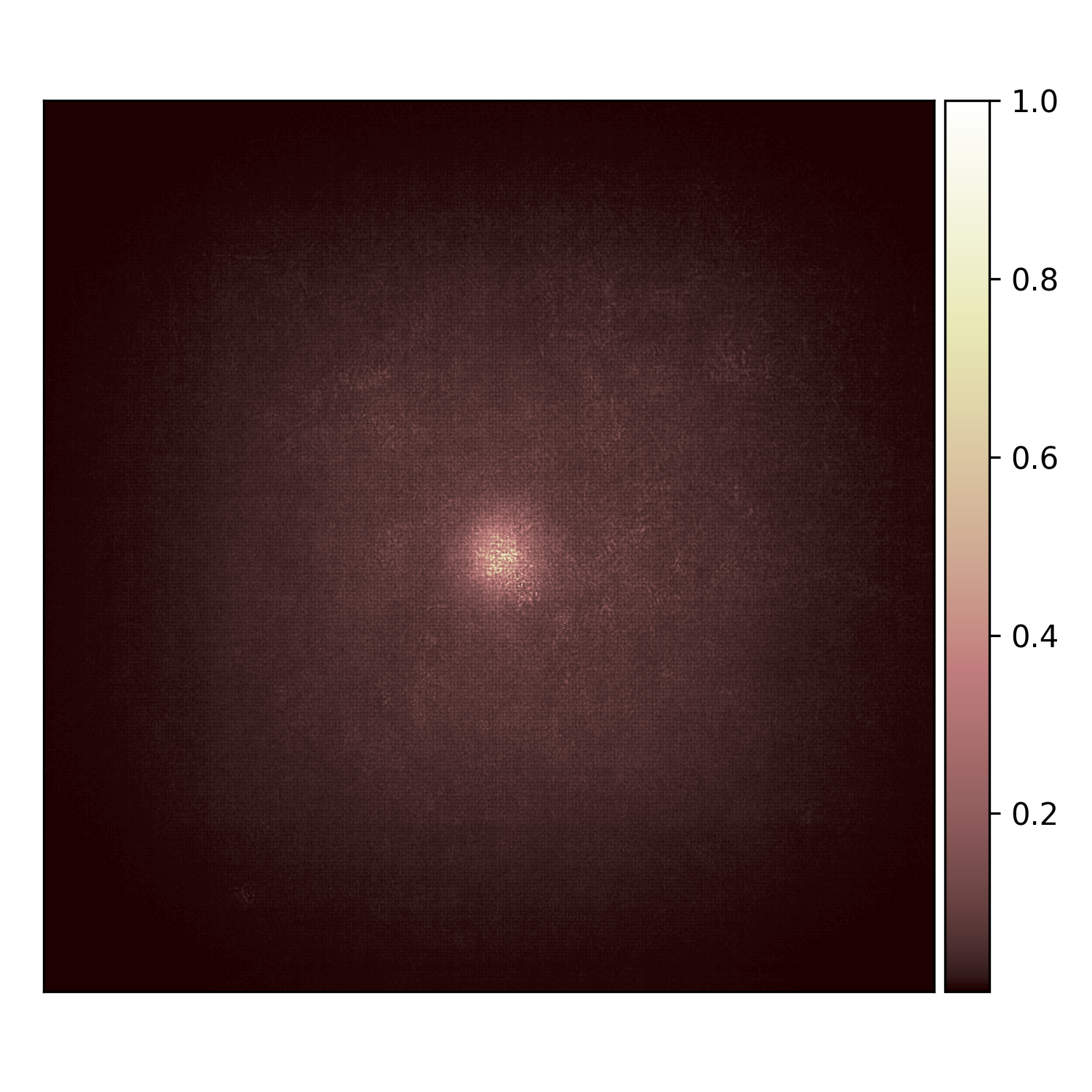}
              \label{fig:erf_k53}
            }
        \subfloat[VAN-LSKA-Tiny (k=65)]{\includegraphics[width=0.25\linewidth,height=0.25\linewidth]{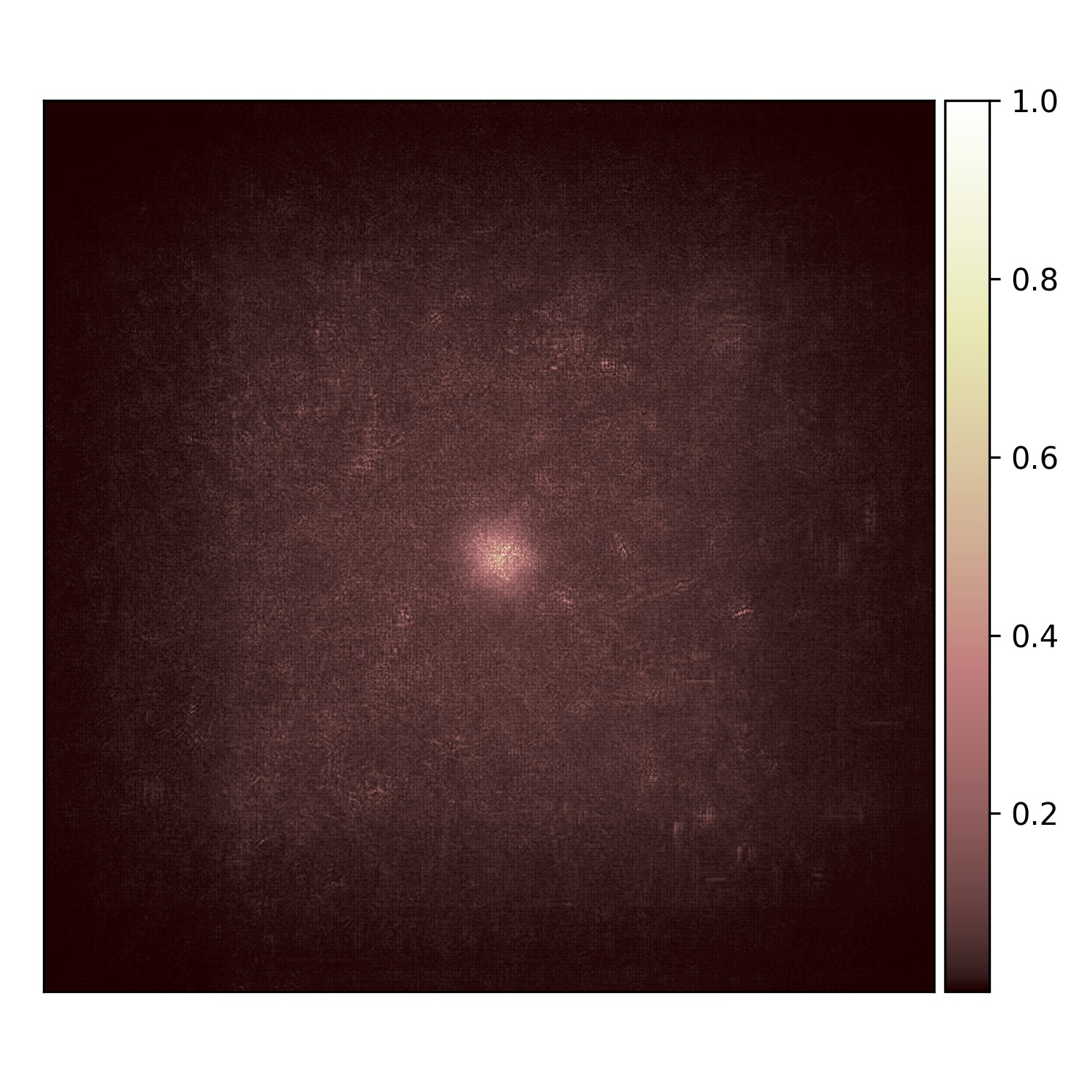}
              \label{fig:erf_k65}
            }
      \caption{The Effective Receptive Field (ERF) of VAN-LSKA-Tiny with kernel size (a) 7-7-7-7 (b) 11-11-11-11 (c) 23-23-23-23 (d) 35-35-35-35 (e) 53-53-53-53 (f) 65-65-65-65. A greater extent of dark areas distributed across the heatmap indicates a larger ERF. VAN-LSKA can capture long-range dependence by enlarging kernel size from 7 to 53.}
      \label{fig:erf}
\end{figure*}


\begin{table}[!t]
        \caption{Properties of convolution (Conv), self-attention, large kernel attention (LKA), and large separable kernel attention (LSKA). $n$ represents the number of pixels.}
        \centering
        \resizebox{1\columnwidth}{!}{
        \begin{tabular}{l|c|c|c|c}
            \hline
            \textbf{Properties}  & \textbf{Conv} &	\textbf{Self-attention} & 	\textbf{LKA} & \textbf{LSKA} \\
            \hline
            Long-range dependence & \xmark & \cmark & \cmark & \cmark \\
            \hline
            Spatial Adaptability & \xmark & \cmark & \cmark & \cmark \\ 
            \hline
            Channel Adaptability & \xmark & \xmark & \cmark & \cmark \\ 
            \hline
            Scalability to extreme large kernel  & \xmark & \xmark & \xmark & \cmark \\
            \hline
            Computational complexity  & $O(n)$ & $O(n^2)$ & $O(n)$ & $O(n)$ \\
            \hline
        \end{tabular}
        }
    \label{tab:lska-prop}
\end{table}


\subsection{Properties of LSKA}
By revisiting the previous attention mechanisms \cite{hu2018squeeze, wang2018non,dosovitskiy2020vit,liu2021swin}, there are four important properties that make LSKA module successful compared to generic convolution, self-attention, and LKA modules as shown in Table \ref{tab:lska-prop}. 

\textbf{Long-range dependence.} As mentioned in Section \ref{sec:Attention}, the self-attention mechanism \cite{vaswani2017attention} is a key component that enables the transformers to model the long-range dependencies. However, recent works \cite{liu2022convnet,ding2022scaling} demonstrate large kernel is an alternative way to capture global information. To achieve this goal, we follow the design of LKA in \cite{guo2022visual}, which decomposes a large kernel into two small kernels, instead of using the trivial large kernel design in \cite{liu2022convnet,ding2022scaling} due to its high computational footprint and optimization difficulty \cite{peng2017large}. To validate the long-range dependence of our proposed LSKA, we utilize the Effective Receptive Field (ERF) generation methodology described in \cite{luo2016understanding, ding2022scaling, liu2023more} to generate the ERF plots of VAN-LSKA-Tiny for kernel sizes ranging from 7 to 65 as shown in figure \ref{fig:erf}. A greater extent of dark areas distributed across the heatmap indicates a larger ERF. From figure \ref{fig:erf_k7} to \ref{fig:erf_k65}, we observe that the dark region spread from kernel size 7 to 65, indicating that the LSKA method can effectively capture long-range dependence in the images.

\textbf{Spatial and channel adaptability.} Spatial attention and channel attention are two common strategies that adaptively recalibrate the weights of the features based on contextual dependencies as mentioned in the Section \ref{sec:Attention}. Our work inherits the design of LKA that includes both properties with lower parameters and computational complexity compared to self-attention. 
The difference between LKA and LSKA is that we adopted a cascaded horizontal and vertical kernel to further reduce the memory and computational complexity as shown in Fig. \ref{fig:LKA-vs-LSKA-d}. 

\textbf{Scalability to the extreme large kernel.} As shown in Fig. \ref{fig:MemComplex}, the LKA-trivial in VAN incurs a quadratic increase in the computational cost with increasing kernel size. The LKA design significantly reduces the computational footprints by a large margin however the number of model parameters grows by increasing the kernel size beyond $23\times23$. 
When the recent state-of-the-art SLaK-Decomposed and SLaK-Sparse-Decomposed methods \cite{liu2023more} are introduced in the VAN, they incur a lower parameters count and computational footprint than LKA, when the kernel size exceeds 100. Note that the results in Fig.\ref{fig:MemComplex} are reported for the VAN-Small network.
Surprisingly, the proposed LSKA-trivial and LSKA versions of LKA-trivial and LKA respectively not only reduce the computational cost but also maintains the number of model parameters of VAN comparatively constant compared to LKA and SLaK. Note that the kernel size also refers to the Maximum Receptive Field (MRF). 

Regarding accuracy performance, as presented in Table \ref{tab:performance-comp-base}, LSKA-Base demonstrates continuous growth as the kernel size increases from 23 to 53. Conversely, LKA-Base starts to saturate beyond a kernel size of 23. These results indicate that LSKA exhibits scalability to extremely large kernels in terms of parameter size, GFLOPs, and accuracy.

\subsection{Complexity Analysis of LSKA}
In this subsection, we compute the number of Floating-Point Operations (FLOPs) and parameters of the proposed LSKA-trivial, LSKA, LKA-trivial, and LKA attention modules as shown in Fig. \ref{fig:MemComplex}. Note that the bias term is ignored in the following analysis for simplifying the calculation. We also assume that the input size and output size of the feature map to the LSKA and LKA are the same (i.e., $H\times W\times C$). For the sake of brevity, we only provide the equations to compute the FLOPs and parameters for LSKA and LKA. However, the same equations can be used to compute the parameters and FLOPs of LKA-trivial and LSKA-trivial. The parameters and FLOPs of the original LKA can be calculated as follows:
\begin{equation}
   Param=(2d-1)^2\times C+\lfloor\frac{k}{d}\rfloor^2\times C+C\times C
    \label{equ:e7}
\end{equation}
\begin{equation}
   FLOPs = ((2d-1)^2\times C+\lfloor\frac{k}{d}\rfloor^2\times C+C\times C)\times H\times W
    \label{equ:e8}
\end{equation}

where ${k}$ is the kernel size and $d$ is the dilation rate. The total number of FLOPs and parameters of the LSKA attention module can be calculated as follows:
\begin{equation}
    Param=(2d-1) \times C \times 2+ \lfloor\frac{k}{d}\rfloor\times C\times2+C\times C
    \label{equ:e5}
\end{equation}
\begin{equation}
FLOPs=((2d-1)\times C\times2+\lfloor\frac{k}{d}\rfloor\times C\times2+C\times C)\times H\times W
    \label{equ:e6}
\end{equation}

By equating the first term of Eq. \ref{equ:e5} and \ref{equ:e7}, we note that the proposed LSKA can save $\frac{2d-1}{2}$ parameters in the depth-wise convolutional layer of the original LKA design. Similarly, by comparing the second terms of Eq. \ref{equ:e5} and \ref{equ:e7}, we note that the proposed LSKA can save $\frac{1}{2}\left\lfloor\frac{k}{d}\right\rfloor$ number of parameters in the dilated-depth-wise convolutional layer of the original LKA design. The saving in terms of the number of flops is the same as the parameters. One can also see that LSKA is computationally more effective than LSKA-trivial. Therefore, unless otherwise specified, we report the performance of LSKA when making the comparison with LKA and the state-of-the-art methods.  
\begin{table}[t]
    \caption{Detail settings of VAN-LKSA. The overall architecture follows VAN \cite{guo2022visual}.}
    \centering
    \resizebox{1\columnwidth}{!}{
    \begin{tabular}{l|c|c|c|c|c}
        \hline
        \textbf{} &
        \textbf{Output Size} &
        \textbf{Layer name} &
        \textbf{VAN-LSKA-Tiny} &
        \textbf{VAN-LSKA-Small} &
        \textbf{VAN-LSKA-Base}
        \\
        \hline
        \multirow{4}{*}{Stage 1} & \multirow{4}{*}{$\frac{H}{4} \times \frac{W}{4}$} & stem  & \multicolumn{3}{c}{$S_1=4; K_1=7$} \\
        \cline{3-6}
        & & Convolution Encoder & 
        \begin{tabular}{@{}c@{}}$C_1=32$ \\ $L_1=3$ \\ $E_1=8$\end{tabular} &
        \begin{tabular}{@{}c@{}}$C_1=64$ \\ $L_1=2$ \\ $E_1=8$\end{tabular} &
        \begin{tabular}{@{}c@{}}$C_1=64$ \\ $L_1=3$ \\ $E_1=8$\end{tabular}
        \\
        \hline
        \multirow{4}{*}{Stage 2} & \multirow{4}{*}{$\frac{H}{8} \times \frac{W}{8}$} & Downsampling  & \multicolumn{3}{c}{$S_1=2; K_1=3$} \\
        \cline{3-6}
        & & Convolution Encoder & 
        \begin{tabular}{@{}c@{}}$C_2=64$ \\ $L_2=3$ \\ $E_2=8$\end{tabular} &
        \begin{tabular}{@{}c@{}}$C_2=128$ \\ $L_2=2$ \\ $E_2=8$\end{tabular} &
        \begin{tabular}{@{}c@{}}$C_2=128$ \\ $L_2=3$ \\ $E_2=8$\end{tabular}
        \\
        \hline
        \multirow{4}{*}{Stage 3} & \multirow{4}{*}{$\frac{H}{16} \times \frac{W}{16}$} & Downsampling & \multicolumn{3}{c}{$S_1=2; K_1=3$} \\
        \cline{3-6}
        & & Convolution Encoder & 
        \begin{tabular}{@{}c@{}}$C_3=160$ \\ $L_3=5$ \\ $E_3=4$\end{tabular} &
        \begin{tabular}{@{}c@{}}$C_3=320$ \\ $L_3=4$ \\ $E_3=4$\end{tabular} &
        \begin{tabular}{@{}c@{}}$C_3=320$ \\ $L_3=12$ \\ $E_3=4$\end{tabular}
        \\
        \hline
        \multirow{4}{*}{Stage 4} & \multirow{4}{*}{$\frac{H}{32} \times \frac{W}{32}$} & Downsampling & \multicolumn{3}{c}{$S_1=2; K_1=3$} \\
        \cline{3-6}
        & & Convolution Encoder & 
        \begin{tabular}{@{}c@{}}$C_4=256$ \\ $L_4=2$ \\ $E_4=4$\end{tabular} &
        \begin{tabular}{@{}c@{}}$C_4=512$ \\ $L_4=2$ \\ $E_4=4$\end{tabular} &
        \begin{tabular}{@{}c@{}}$C_4=512$ \\ $L_4=3$ \\ $E_4=4$\end{tabular}
        \\
        \hline
        \end{tabular}
    }
    \label{tab:model-instantiation}
\end{table}

\begin{figure}[t]
\centering
\includegraphics[width=\linewidth]{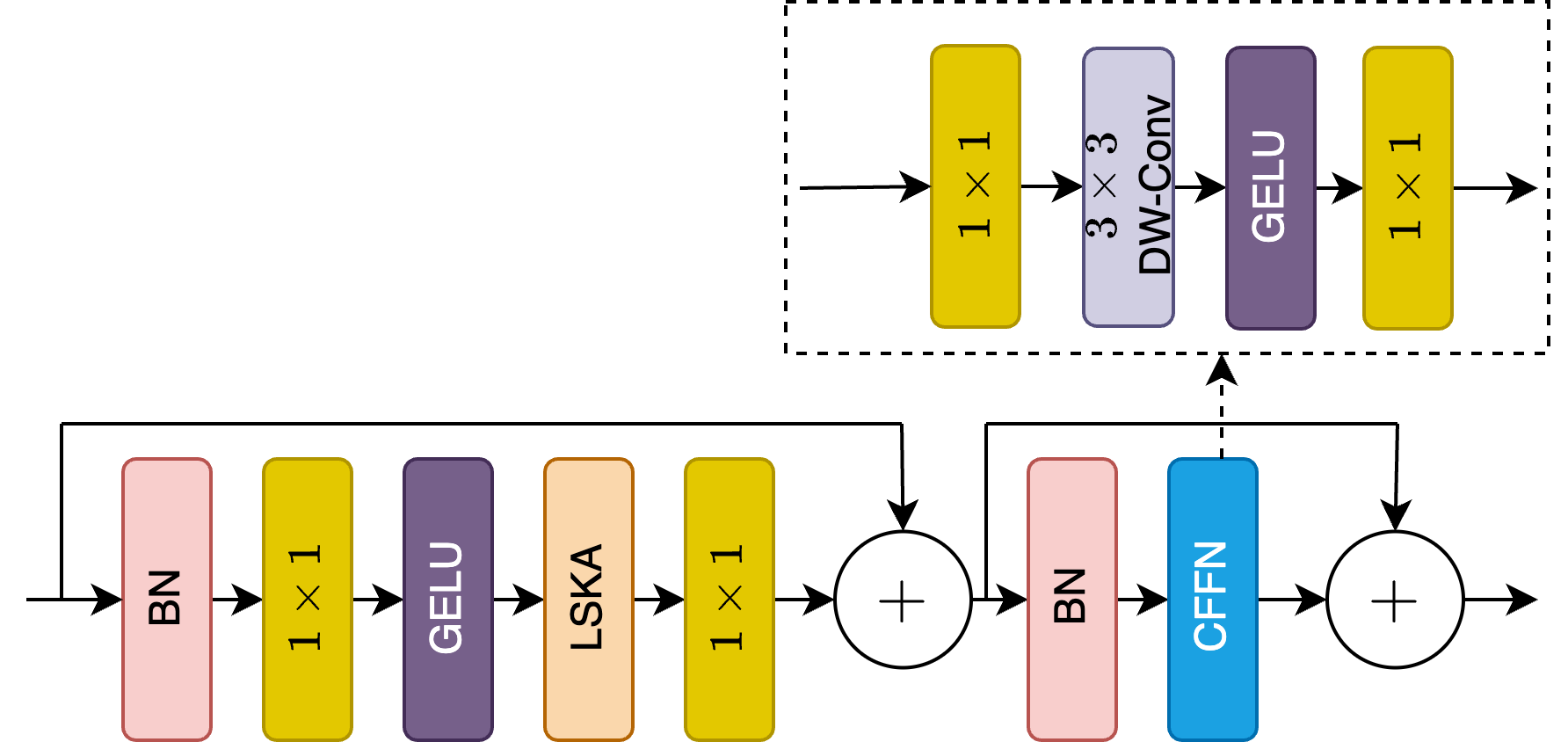}
\caption{Block design of VAN with LSKA. The design follows VAN \cite{guo2022visual} and the difference is that we replace LKA with our proposed LSKA. BN, GELU, and CFFN refer to batch normalization \cite{ioffe2015batch}, gaussian error linear unit \cite{hendrycks2016gaussian} and convolutional feed-forward layer \cite{wang2022pvt} respectively.}
\label{fig:VANStage}
\end{figure}
\subsection{Model Architecture}
In this work, we follow the architecture design of VAN \cite{guo2022visual} as shown in Table \ref{tab:model-instantiation}. The hyperparameters of the model are listed as follows:
\begin{itemize}
\item {${S_i}$: the stride of the convolution layer in the input stem and downsampling in stage $i$;}
\item {${K_i}$: the kernel size of the convolution layer in input stem and downsampling in stage $i$;}
\item {${C_i}$: the number of output channels in stage $i$;}
\item {${E_i}$: the expansion ratio of convolutional feed-forward layer in stage $i$;}
\item {${L_i}$: the number of block in stage $i$;}
\end{itemize}
Following the design in VAN, our model consists of an input stem layer and four subsequent stages. The first layer of the input stem contains a $7\times 7$ convolution layer with stride 4 followed by a batch normalization layer. This layer reduces the input resolution by 4 and increases the channel to 32 or 64 depending on the model capacity.  

Except for stage 1, each stage begins with a $3\times3$ convolution with stride 2 downsampling layer. Then it is followed by a convolution block which contains a batch normalization, LSKA module and a convolution feed-forward network (CFFN) \cite{wang2022pvt} as shown in Fig. \ref{fig:VANStage}. As a common practice in \cite{sandler2018mobilenetv2, xie2017aggregated, liu2022convnet, ding2022scaling}, our model includes a $1\times1$ convolution layer before and after the depth-wise convolution for channel interaction. To provide more non-linearities, GELU \cite{hendrycks2016gaussian} activation layers are attached before LSKA and inside the CFFN. The major difference between the LKA in VAN and our work is that we replace the LKA layer with our LSKA layer for every convolution block. 

To provide more instances of discussion, we design three different capacities of VAN with LSKA, namely VAN-LSKA-Tiny, VAN-LSKA-Small, and VAN-LSKA-Base. These models contain the same number of convolution blocks, the number of channels, and the expansion ratio of CFFN as VAN with LKA for a fair comparison between both attention modules in Section \ref{sec:experiment}.

\section{Experiment}
\label{sec:experiment}
To verify the effectiveness of our proposed large separable kernel attention (LSKA), we compare the performance of VAN with LSKA against the baseline VAN with LKA \cite{guo2022visual} on image classification, object detection, semantic segmentation, and robustness evaluation. It should be noted that the seminal work of the VAN network did not provide any robustness evaluation on the different versions of perturbation in the input data. Additionally, we also benchmark the robustness evaluation of LKA-based VAN, LSKA-based VAN, ConvNeXt \cite{liu2022convnet}, and  the state-of-the-art ViTs \cite{wang2021pyramid, liu2021swin, touvron2021training, wang2022pvt}.

\begin{table}[t]
    \centering
    \caption{Comparisons with VAN-LKA baseline on ImageNet-1K classification. Params represent the number of model parameters. GFLOPs represent floating point operations. Top-1 represents Top-1 accuracy. The inference speed is the number of samples per second (Higher is better). Here "GFLOPs" is calculated with input size $224 \times 224$.}
    \resizebox{1\columnwidth}{!}{
    \begin{tabular}{l|c|c|c|c|c}
        \hline
        \textbf{Model}  & \textbf{Kernel size ($k$)} & \textbf{Params.(M)} &	\textbf{GFLOPs} & \textbf{Inference Speed} & \textbf{Top-1} \\
        \hline
        VAN-LKA-Tiny & 23-23-23-23 & 4.1  & 0.9 & 660 & 74.7   \\
        VAN-LSKA-Tiny (our) & 23-23-23-23 & \textbf{4.0} & \textbf{0.8} & \textbf{713} & \textbf{75.0} \\
        \hline
        VAN-LKA-Small & 23-23-23-23 & 13.9 & 2.5 & 430 & 80.5 \\
        VAN-LSKA-Small (our) & 23-23-23-23 & \textbf{13.7} & \textbf{2.4} & \textbf{454} & \textbf{80.6} \\
        \hline
        VAN-LKA-Base & 23-23-23-23 & 26.6 & 5.0 & 227 & 82.8 \\
        VAN-LSKA-Base (our) & 23-23-23-23 & \textbf{26.2} & \textbf{4.9} & \textbf{238} & \textbf{82.8} \\
        \hline
        \end{tabular}
    }
    \label{tab:lka-vs-lska-classification}
\end{table}
\subsection{Image Classification}
\label{sec:image-classification}
\textbf{Settings.} We conduct the image classification experiments on ImageNet-1K dataset \cite{deng2009imagenet} which consists of 1.2M training images with 1000 categories and 50K validation images. All models were trained in the same settings for a fair comparison and we report the top-1 accuracy on the validation set. Following the VAN-LKA training settings as the baseline, a $224\times224$ training sample is randomly cropped from an image. We adopted random horizontal flipping, label smoothing \cite{muller2019does}, mixup \cite{zhang2018mixup}, cutmix \cite{yun2019cutmix}, and random erasing \cite{zhong2020random} for the data augmentation. During the training, we use AdamW \cite{loshchilov2018decoupled} with a momentum of 0.9, weight decay $5\times10^{-2}$, and a batch size of 64. We train the models for 5 warm-up epochs with a learning rate of $1\times10^{-6}$ followed by a normal 300 epochs training with an initial learning rate of $5\times10^{-4}$ and decreasing with a cosine decaying schedule. We also employ a variant of LayerScale \cite{guo2022visual} in the attention layer and feed-forward layer with an initial diagonal value of 0.01. 

For a fair comparison with LKA module in VAN, we use the same kernel size and dilation rate setting (i.e., $k=23$ and $d=3$) as VAN. During the testing stage, we apply the center crop with image size $224\times224$ on the validation set and report the top-1 accuracy, the number of parameters, and inference speed (example/secs). The inference speed of all models is tested with batch size 64 and image size $224\times224$ on a 1080Ti GPU. We first feed 50 batches to warm up the hardware followed by 50 batches to record the average running time. 

\textbf{Results.} In Table \ref{tab:lka-vs-lska-classification}, we show that our LSKA achieves a lower number of parameters and computational flops while maintaining a similar performance of LKA in VAN. For example, VAN-LSKA-Tiny saves 2.4\% of parameters and 11.1\% of computational flops with a slightly better performance compared to VAN-LKA-Tiny. VAN-LSKA-Small has a 1.4\% and 4\% lower parameters and computational flops with comparable accuracy compared to VAN-LKA-Small. On the other hand, VAN-LSKA-Base achieves a 1.5\% and 2\% lower parameter and computational flops respectively, while it maintains the same accuracy as VAN-LKA-Base. In section \ref{sssec:ab-studies-diff-decomp}, we further demonstrate that LSKA achieves a better speed-accuracy trade-off than LKA.
\begin{table*}[t]
    \caption{Object detection and instance segmentation performance of VAN-LKA and VAN-LSKA with different capacity of backbone on COCO2017 dataset. We follow the same training and validation framework MMDetection in Swin Transformer. ${\rm AP}^b$ and ${\rm AP}^m$ represent the bounding box AP and mask AP respectively. Here "GFLOPs" is calculated with an input size $1280\times800$.}
    \centering
    \resizebox{2\columnwidth}{!}{
    \begin{tabular}{l|c|c|c|c|c|c|c|c|c|c|c|c|c|c|c|c|c}
        \hline
        \multirow{2}{*}{Backbone} & \multirow{2}{*}{Kernel size ($k$)} & \multicolumn{8}{c}{Mask R-CNN 1x} & \multicolumn{7}{|c}{RetinaNet 1x} \\ 
        \cline{3-18}
        & & Params(M) & GFLOPs & ${\rm AP}^b$ & ${\rm AP}_{50}^b$ & ${\rm AP}_{75}^b$ & ${\rm AP}^m$  & ${\rm AP}_{50}^m$ & ${\rm AP}_{75}^m$ & Params(M) & GFLOPs & {\rm AP} & ${\rm AP}_{50}$ & ${\rm AP}_{75}$ & ${\rm AP}_{S}$  & ${\rm AP}_{M}$ & ${\rm AP}_{L}$ \\
        \hline
        VAN-LKA-Tiny & 23-23-23-23 & 23.9 & 187.2 & \textbf{41.1}  & \textbf{62.8}   &  \textbf{44.6}  &  \textbf{37.8}  &  \textbf{59.8}  &  \textbf{40.4}  & 13.4 & 169.0  & \textbf{40.1}  & \textbf{60.2}  & \textbf{43.1}  & \textbf{24.2} &  \textbf{44.2}  & \textbf{52.8}  \\
        VAN-LSKA-Tiny & 23-23-23-23 & \textbf{23.8}  & \textbf{186.6} &  40.9  &  62.4  & 44.5   & 37.3   & 59.2   &   39.8 & \textbf{13.3}  & \textbf{168.3} & 39.7 &  59.7  &  42.1  &  23.6  & 43.9 & 51.3\\
        \hline
        VAN-LKA-Small &  23-23-23-23 & 33.5 & 221.6 &  43.9 &  \textbf{65.8}  & \textbf{48.0}   & 40.0  & \textbf{62.9}   &  43.0 & 23.6  & 203.0 & 43.2  & 63.9 & 46.0 &  27.5 &  47.3  & 55.2  \\
        VAN-LSKA-Small & 23-23-23-23 & \textbf{33.4} & \textbf{220.7}  &  \textbf{44.0} &  65.6  & 47.9   & \textbf{40.0}  & 62.7  & \textbf{43.0} & \textbf{23.5}  &  \textbf{202.1} & \textbf{43.3} & \textbf{64.0}  & \textbf{46.5} & \textbf{29.2}  & \textbf{47.4}  & \textbf{56.0} \\
        \hline
        VAN-LKA-Base &  23-23-23-23 & 46.2 & 272.9 &  \textbf{46.6}  & \textbf{68.2}  & \textbf{50.9} & \textbf{41.9} &  \textbf{65.1} & \textbf{45.0}  & 36.3  & 254.3 & \textbf{45.0}  & \textbf{65.9}  &  \textbf{48.2} &  \textbf{27.8} &  \textbf{49.1}  &  \textbf{58.5} \\
        VAN-LSKA-Base & 23-23-23-23 &  \textbf{45.9} & \textbf{271.2} & 46.3 &  67.8 & 50.7 & 41.5 & 64.7 & 44.6 & \textbf{36.0}  &  \textbf{252.6} &  44.5 & 65.0 & 47.8  &  27.6 & 48.6 & 57.9\\
        \hline
        \end{tabular}
    }
    
    \label{tab:lka-vs-lska-obj-detection}
\end{table*}
\subsection{Object Detection on COCO}
\label{sec:object-detection}
\textbf{Settings.} We conduct the object detection and instance segmentation on COCO 2017 benchmark \cite{lin2014microsoft} which consists of 118K images in the training dataset and 5K images in the validation set. We compare both VAN-LKA and VAN-LSKA as the backbones for two standard detectors, the Mask R-CNN \cite{he2017mask} and the RetinaNet \cite{lin2017focal}. For a fair comparison, all models adopt the same training settings as reported in \cite{guo2022visual}. Following the common practice, before the training, we initialize the backbone and newly added layers by using the pre-trained weights on ImageNet and Xavier \cite{glorot2010understanding} respectively. During the training, we use AdamW optimizer with an initial learning rate $1\times10^{-4}$, weight decay $5\times10^{-2}$, batch size 16 and 1$\times$ schedule (i.e., 12 epochs). The shorter size of the training image is resized to 800, while the long side is at most 1333. During the testing phase, we use the same image rescaling scheme as the training stage before feeding into the network. All implementations are based on MMDetection \cite{mmdetection}.

\textbf{Results.} As shown in Table \ref{tab:lka-vs-lska-obj-detection},  the proposed LSKA design as a backbone in Mask R-CNN and RetinaNet results in a lower number of model parameters and model FLOPS while providing comparatively similar performance as LKA. For example, when we use Mask R-CNN as an object detector, VAN-LSKA-Base saves 0.3 million parameters and 2 GFLOPs of computation while achieving a comparable performance with only a 0.3\% decrease compared to VAN-LKA-Base (46.3 vs 46.6). When we change the object detector to RetinaNet, VAN-LSKA-Base saves 0.3 million parameters and 1 GFLOPs of computation with similar performance compared to VAN-LKA-Base (45.0 vs 44.5). Similar results are found in other model capacities. For example, when we use Mask R-CNN, VAN-LSKA-Small has 33.4 million parameters and 221 GFLOPs computation time, which is 0.1 million and 1 GFLOPs more than the VAN-LKA-Small. 

\subsection{Semantic Segmentation}
\label{sec:semantic-segmentation}
\textbf{Settings.} Semantic Segmentation experiments are conducted on ADE20K \cite{zhou2017scene} which contains 150 fine-grained semantic categories, 20K training images, 2K validation images, and 3K testing images. We compare our VAN-LSKA and VAN-LKA backbone for the Semantic FPN \cite{kirillov2019panoptic}. Before the training, we initialize the backbone with pre-trained weights on ImageNet and the remaining new layers are initialized with Xavier. Following the common practice, \cite{wang2021pyramid, kirillov2019panoptic, guo2022visual, wang2022pvt}, during the training, we use AdamW optimizer with an initial learning rate $1\times10^{-4}$, weight decay $1\times10^{-4}$, batch size 16, and 40K iterations. The learning rate is decayed following the polynomial decay function with power 0.9. All the training images are resized to $2048\times512$ with a ratio range of 0.5 to 2.0 followed by randomly cropping them with a size of $512\times512$. During the testing, the images are resized to $512$ along the shorter side and the longer size is resized to at most $2048$. All implementations are based on MMSegmentation \cite{mmseg2020}.

\begin{table}[t]
    \caption{Semantic Segmentation performance of VAN-LKA and VAN-LSKA with different size of model capacity on ADE20K validation set. Here "GFLOPs" is calculated with input size $2048\times1024$.}
    \centering
    \resizebox{1\columnwidth}{!}{
    \begin{tabular}{l|c|c|c|c}
        \hline
        \multirow{2}{*}{Backbone} & \multirow{2}{*}{Kernel size ($k$)} & \multicolumn{3}{c}{Semantic FPN}\\ 
        \cline{3-5}
        & & Params(M) & GFLOPs & mIoU ($\%$) \\
        \hline
        VAN-LKA-Tiny & 23-23-23-23 & 8.0 & 206  & \textbf{40.0} \\
        VAN-LSKA-Tiny & 23-23-23-23 & \textbf{7.9} & \textbf{205} & 39.5 \\
        \hline
        VAN-LKA-Small & 23-23-23-23 & 17.6 & 277  & \textbf{43.3} \\
        VAN-LSKA-Small & 23-23-23-23 & \textbf{17.5} & \textbf{275} & 42.6 \\
        \hline
        VAN-LKA-Base & 23-23-23-23 & 30.3 & 382 & 45.8 \\
        VAN-LSKA-Base & 23-23-23-23 & \textbf{30.1} & \textbf{378} & \textbf{46.1} \\
        \hline
        \end{tabular}
    }
    
    \label{tab:lka-vs-lska-segmentation}
\end{table}

\textbf{Results.} As shown in Table \ref{tab:lka-vs-lska-segmentation}, when using Semantic FPN for semantic segmentation, we found that the results are consistent with image classification and object detection. For example, VAN-LSKA-Base saves 0.2M parameters and reduces the GLOPS by $1.05\%$ while obtaining a slightly better performance compared to VAN-LKA-Base. Similarly, we found that VAN-LSKA-Tiny and VAN-LSKA small save 0.1M parameters, reducing the GFLOPS by $0.49\%$ and $0.36\%$ while obtaining comparable performance to VAN-LKA-Tiny and VAN-LKA-Small, respectively.  

\subsection{Robustness Evaluation}
\label{sec:robustness-eval}
\textbf{Settings.} Robustness experiments are conducted on five diverse ImageNet datasets with different perspectives including common corruption, natural adversarial examples, semantic shift, out-of-domain distribution, and perceptual dissimilar from common corruption. Table \ref{tab:robust-dataset-summary} shows the summary of the datasets and their corresponding purpose. In the experiment, we report the top-1 accuracy on ImageNet-C \cite{hendrycks2019benchmarking}, ImageNet-A \cite{hendrycks2021natural}, ImageNet-R \cite{hendrycks2021many}, and ImageNet-SK datasets \cite{wang2019learning}. Following \cite{hendrycks2019benchmarking}, we report the mean corruption error (mCE) (lower values is better) for ImageNet-C and ImageNet-$\mathrm{\bar{C}}$. We also report the retention rate (Ret R) (higher values are better) for ImageNet-C following \cite{zhou2022understanding} to quantify the resistance of a model against corruption. Ret R defines as follows:
\begin{equation}
    \resizebox{.5\hsize}{!}{$Ret R=\dfrac{ImageNet\text{-}C\text{ }Top\text{-}1\text{ }Accuracy}{ImageNet\text{-}1K\text{ }Top\text{-}1\text{ }Accuracy}$}
\end{equation}

\begin{table*}[t]
    \caption{Robustness comparison among VAN-LKA, and VAN-LSKA with different model capacities on ImageNet-C (IN-C), ImageNet-$\mathrm{\bar{C}}$ (IN-$\mathrm{\bar{C}}$), ImageNet-A (IN-A), ImageNet-R (IN-R) and ImageNet-Sketch (IN-SK). We report top-1 accuracy, mean corruption error (mCE), and retention rate (Ret R). For ImageNet-$\mathrm{\bar{C}}$., we report corruption error (IN-$\mathrm{\bar{C}}$). $\uparrow$ means higher is better, whereas $\downarrow$ means lower is better. Here, GFLOPs is calculated with input size $224 \times 224$.}
    \centering
    \resizebox{2\columnwidth}{!}{
   
    \begin{tabular}{lcccc|ccc|cccc}
    \hline
        \multicolumn{12}{c}{}\\
        \textbf{Model} &
        \textbf{Kernel size} &
        \textbf{Param (M)} &
        \textbf{GFLOPs} &
        \textbf{IN ($\uparrow$)} &
        \textbf{IN-C ($\uparrow$)} &
        \textbf{mCE ($\downarrow$)} &
        \textbf{Ret R ($\uparrow$)} &
        \textbf{IN-$\mathrm{\bar{\textbf{C}}}$ ($\downarrow$)} &
        \textbf{IN-A ($\uparrow$)} & 
        \textbf{IN-R ($\uparrow$)} &
        \textbf{IN-SK ($\uparrow$)}
        \\
        \hline
        VAN-LKA-Tiny & 23-23-23-23 & 4.1 & 0.9  &  74.7 & 46.0 & 69.0 & 61.5 & \textbf{50.5} & 7.1 & \textbf{39.1} & \textbf{26.2}\\
        VAN-LSKA-Tiny & 23-23-23-23 & \textbf{4.0} & \textbf{0.8}  & \textbf{75.0} & \textbf{46.3} & \textbf{68.6} & \textbf{61.9} & 50.6 & \textbf{7.5} & 38.5 & 25.7 \\
        \hline
        VAN-LKA-Small & 23-23-23-23  & 13.9 & 2.5  & 80.5 & \textbf{55.2} & \textbf{57.4} & \textbf{68.5} & 42.7 & 18.0 & 44.4 & 31.7 \\
        VAN-LSKA-Small &  23-23-23-23  & \textbf{13.7} & \textbf{2.4} & \textbf{80.6} & 54.9 & 57.6 & 68.1 & \textbf{42.5} & \textbf{19.3} & \textbf{44.4} & \textbf{31.7}\\
        \hline
        VAN-LKA-Base & 23-23-23-23  & 26.6 &  5.0 & 82.8 & \textbf{60.2} & \textbf{51.1} & \textbf{72.7} & \textbf{37.5} & \textbf{31.1} & 46.8 & 34.8 \\
        VAN-LSKA-Base &  23-23-23-23  & \textbf{26.2} & \textbf{4.9}  & \textbf{82.8} & 60.1 & 51.2 & 72.6 & 37.7 & 29.6 & \textbf{46.9} & \textbf{35.0} \\
        \hline
        \end{tabular}
        }
        
        \label{tab:lka-vs-lska-robustness}
\end{table*}

\begin{table}[t]
    \caption{Summary of the robustness evaluation datasets and their properties}
    \centering
    \resizebox{1\columnwidth}{!}{
    \begin{tabular}{l|c}
        \hline
        \textbf{Dataset} &
        \textbf{Purpose}
        \\
        \hline
        ImageNet-C \cite{hendrycks2019benchmarking} & Common corruptions\\
        \hline
        \vspace{-0.25cm}
        &  \\
        ImageNet-$\mathrm{\bar{C}}$ \cite{mintun2021interaction} & Perceptual dissimilar from common corruption\\
        \hline
        ImageNet-A \cite{hendrycks2021natural} & Natural adversarial examples\\
        \hline
        ImageNet-R \cite{hendrycks2021many} & Semantic shift\\
        \hline
        ImageNet-SK \cite{wang2019learning} & Out-of-domain distribution\\
        \hline
        \end{tabular}
        }
        
        \label{tab:robust-dataset-summary}
\end{table}

\textbf{Results.} As shown in Table \ref{tab:lka-vs-lska-robustness}, we observe that VAN-LSKA achieves a comparable performance across different parameter regimes with only 0.1\%-0.6\% differences compared to VAN-LKA. At the same time, VAN-LSKA saves 2.5\%, 1.4\%, and 1.5\% parameters in Tiny, Small, and Base models respectively. Meanwhile, VAN-LSKA uses 11.1\%, 4\%, and 2\% less computational time in Tiny, Small, and Base models, respectively. These results are consistent with previous results in image classification, object detection, and semantic segmentation.

\section{Ablation Studies}
\label{sec:ablation-studies}
\subsection{Evaluation of different large kernel decomposition method}
First, we conduct an experiment to explore the effect of different large kernel decomposition methods. Here we compare LKA, LSKA-trivial, and LSKA against LKA-trivial with increasing kernel size. For this purpose, we train the Tiny models on the ImageNet-1K classification task following the setup described in Section \ref{sec:image-classification}.  We vary the kernel size ($k$) unevenly from 7 to 65. We set the dilation rate $d$ as 2, 2, 3, 3, 3, 3 for $k=7, 11, 23, 35, 53, 65$ respectively. The specific setting for the kernel size of LKA-trivial, LSKA-trivial, LKA, and LSKA can be found in Fig. \ref{fig:LKA-vs-LSKA}. It is important to note that the kernel size of LKA and LSKA is the maximum receptive field,  as both modules decompose a large kernel into two relatively small kernels. We report the results for all the methods in terms of top-1 accuracy, model parameters, model FLOPs, and inference speed in Fig.\ref{fig:performance-comp}. We also include the numerical results for Fig.\ref{fig:performance-comp} in the Appendix Table \ref{tab:performance-comp-tiny}.

\begin{figure*}[t]
      \centering
      \subfloat[]{\includegraphics[width=0.4\linewidth]{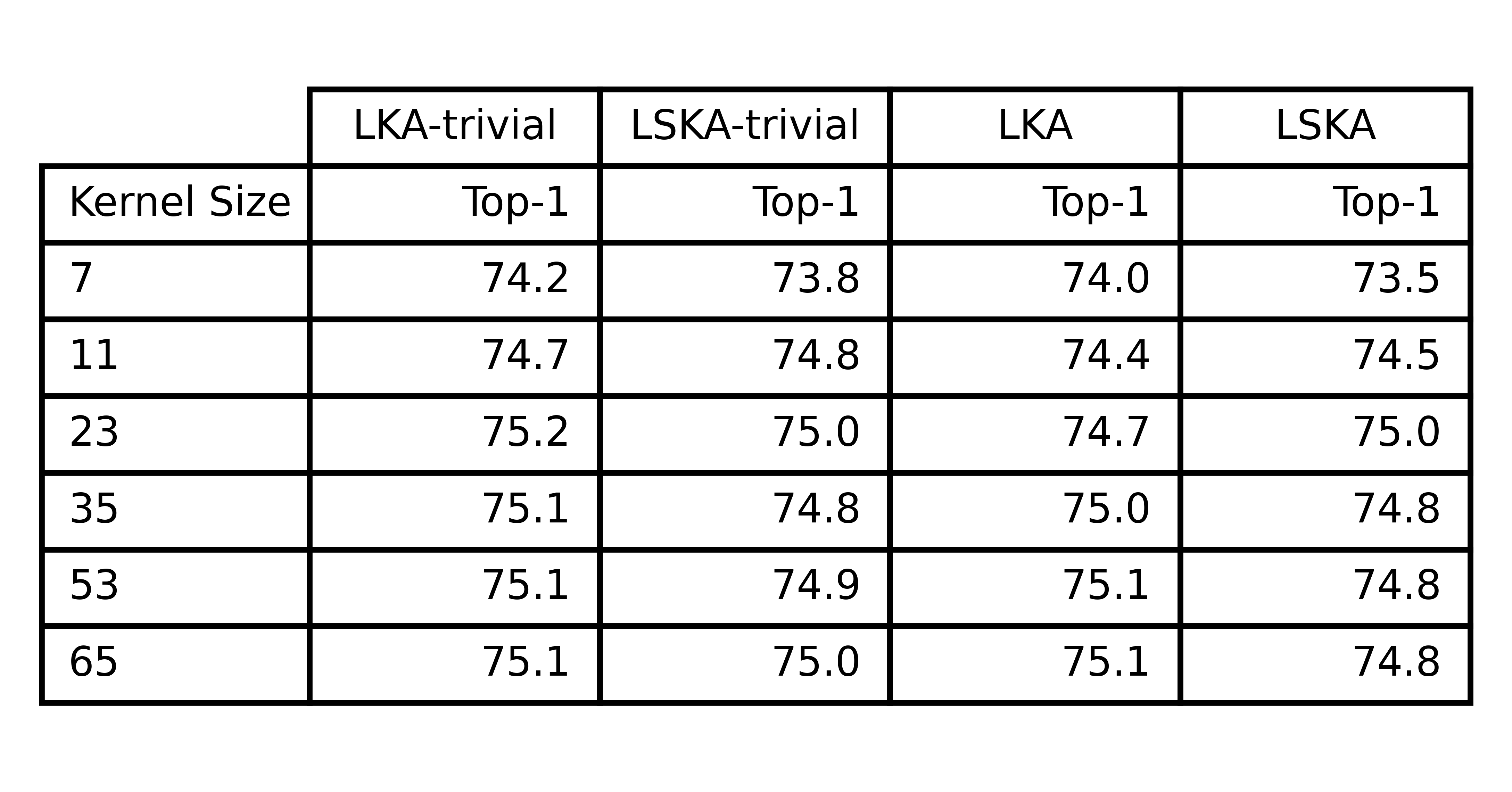}
      \label{fig:acc-vs-kernel-ablation}}
      \subfloat[]{\includegraphics[width=0.4\linewidth]{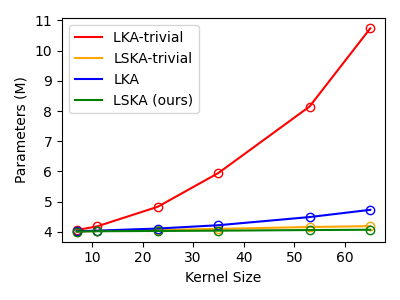}
        \label{fig:param-vs-kernel-ablation}}\\
      \subfloat[]{\includegraphics[width=0.4\linewidth]{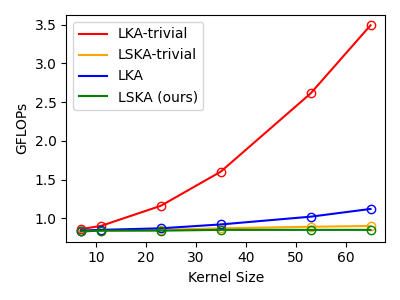}
        \label{fig:gflops-vs-kernel-ablation}}
      \subfloat[]{\includegraphics[width=0.4\linewidth]{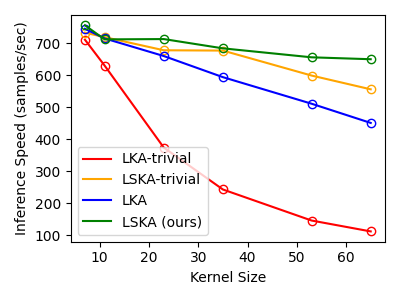}
          \label{fig:inference-vs-kernel-ablation}
        }
      \caption{Comaprsion on the (a). accuracy, (b). model parameters, (c). model FLOPs, and (d). inference speed of LKA-trivial, LKA, LSKA-trivial, and LSKA with different kernel sizes. Note that we use Tiny models to conduct the comparison.}
      \label{fig:performance-comp}
\end{figure*}

\begin{table}[t]
    \caption{Comaprsion among LKA and LSKA with large kernel sizes on Base models. We observe that the performance of LSKA improves as kernel size increases from 7 to 53, while the LKA starts to saturate beyond the kernel size of 23.}
    \centering
    \resizebox{1\columnwidth}{!}{
    \begin{tabular}{l|cccc|cccc}
        \hline
         Backbone & \multicolumn{4}{c|}{LKA} & \multicolumn{4}{c}{LSKA}\\ 
         \hline
         Kernel Size & Params(M) & GFLOPs & Inf. Speed & Top-1 & Params(M) & GFLOPs & Inf. Speed & Top-1 \\
        \hline
        23-23-23-23 & 26.58 & 5.01 & 229 & 82.8 & 26.29 & 4.92 & 237 & 82.8 \\
        35-35-35-35 & 27.01 & 5.13 & 203 & 82.8 & 26.34 & 4.93 & 231 & 82.9 \\
        53-53-53-53 & 28.01 & 5.42 & 179 & 82.8 & 26.41 & 4.95 & 224 & 83.0 \\
        \hline
        \end{tabular}
    }
    \label{tab:performance-comp-base}
\end{table}

One can see in Fig. \ref{fig:performance-comp} that our proposed methods, LSKA-trivial and LSKA, significantly reduced the model parameters and the model FLOPs while achieving comparable performance to LKA-trivial and LKA. One can also see from Fig. \ref{fig:inference-vs-kernel-ablation}, that our proposed LSKA-trivial and LSKA show significantly lower degradation in the inference speed compared to LKA-trivial and LKA when we vary the kernel size from $7\times7$ to $65\times 65$. Moreover, comparing Fig. \ref{fig:acc-vs-kernel-ablation} and Fig. \ref{fig:inference-vs-kernel-ablation}, we can see that LSKA-trivial and LSKA can offer better speed-accuracy trade-offs compared to LKA-trivial and LKA. Compared to LKA-trivial and LKA, LSKA-trivial, and LSKA enable the larger receptive field while maintaining a lower computation time. 

Furthermore, we observe that the performance of LSKA exhibits a slight decrease and begins to saturate as the kernel size is further increased from 23 to 65, as shown in Fig. \ref{fig:acc-vs-kernel-ablation}. We also visualized the effective receptive field of LKA and LSKA to illustrate the performance start to saturate beyond a kernel size of 23. The corresponding plot can be found in Figure \ref{fig:erf-lska-lka} in \ref{erf-comp}. This finding contradicts the results reported in RepLKNet \cite{ding2022scaling} and SLaK \cite{liu2023more}, where a performance improvement was observed when increasing the kernel size from 25 to 31 and 31 to 51, respectively. We conjecture that this saturation is due to insufficient network width and depth. To validate this hypothesis, we conduct a similar experiment on the Base model with LKA and LSKA. As shown in Table \ref{tab:performance-comp-base}, we observe that the LKA reaches a saturation point beyond a kernel size of 23. In contrast, LSKA exhibits improved performance from kernel size 23 to 53. This finding aligns with RepLKNet and SLaK, proving that larger receptive fields enhance long-range dependence. 

\label{sssec:ab-studies-diff-decomp}
\subsection{Varying the kernel size of LSKA and LKA on various downstream tasks}
\label{sssec:ab-studies-diff-kernel-size}
Following the experiments in Section \ref{sec:object-detection}, \ref{sec:semantic-segmentation}, and \ref{sec:robustness-eval}, we explore the effects of increasing the kernel size on the downstream tasks. Before the training for object detection and semantic segmentation, and performing the robustness testing, we initialize the backbone models with the corresponding kernel size pre-trained on ImageNet-1K as mentioned in Section \ref{sssec:ab-studies-diff-decomp}. The other training settings and evaluation methods follow the corresponding tasks mentioned in Section \ref{sec:object-detection}, \ref{sec:semantic-segmentation}, \ref{sec:robustness-eval}. 
\begin{table*}[t]
    \caption{Object detection and instance segmentation performance of VAN-LKA-Tiny and VAN-LSKA-Tiny with different kernel size on COCO2017 dataset. We follow the same training and validation framework MMDetection in Swin Transformer. Here "GFLOPs" is calculated with input size 1280x800.}

    \centering
    \resizebox{2\columnwidth}{!}{
    \begin{tabular}{l|c|c|c|c|c|c|c|c|c|c|c|c|c|c|c|c|c}
        \hline
        \multirow{2}{*}{Backbone} & \multirow{2}{*}{Kernel size($k$)} & \multicolumn{7}{c}{Mask R-CNN 1x} & \multicolumn{7}{|c}{RetinaNet 1x} \\ 
        \cline{3-18}
        & & Params(M) & GFLOPs & ${\rm AP}^b$ & ${\rm AP}_{50}^b$ & ${\rm AP}_{75}^b$ & ${\rm AP}^m$  & ${\rm AP}_{50}^m$ & ${\rm AP}_{75}^m$ & Params(M) & GFLOPs & {\rm AP} & ${\rm AP}_{50}$ & ${\rm AP}_{75}$ & ${\rm AP}_{S}$  & ${\rm AP}_{M}$ & ${\rm AP}_{L}$ \\
        \hline
        \multirow{4}{*}{VAN-LKA-Tiny (Baseline)} & 7 & 23.79 & 186.5 & 39.9  & 61.2   &  43.8  &  36.8  &  58.0  &  39.4 & 13.29 & 168.2 & 38.8  &  58.3  &  41.5  &  22.3  & 42.3 & 50.7 \\
        & 11 & 23.82 & 186.7 & 40.6  &  61.9  &  44.5  &  37.3 & 59.1   & 39.9  & 13.32 & 168.4 & 39.4  & 59.5 & 42.0  & 23.3 &  43.2  & 50.9   \\
        & 23 & 23.88 & 187.2 & 41.1  & 62.8   &  44.6  &  37.8  &  59.8  &  40.4  & 13.38 & 169.0 & 40.1  & 60.2  & \textbf{43.1}  & 24.2 &  44.2  & 52.8  \\
        & 35 & 24.00 & 188.1 & \textbf{41.5}  & \textbf{63.1}   &  \textbf{45.4}  &  \textbf{38.0}  &  \textbf{60.0}  &  \textbf{40.8} & 13.50 & 169.9 & \textbf{40.3}  &  \textbf{60.9}  & 42.6   & \textbf{24.5} &  \textbf{44.4}  & \textbf{52.8}\\
        \hline
        \multirow{4}{*}{VAN-LSKA-Tiny} & 7 & 23.78 & 186.4 & 39.5  & 60.7   &  43.0  &  36.6  &  57.6  &  39.3 & 13.29 & 168.2 & 38.8 &  58.2  & 41.0   & 22.1 &  42.2  & 49.8 \\
        & 11 & 23.79 & 186.5  & 40.3   & 61.8   & 43.7  &  37.2  & 58.8   & 39.7 & 13.30 & 168.2 & 39.2  & 59.2 &  42.0 & 24.1 & 42.7   & 51.3 \\
        & 23 & 23.80  & 186.6 & 40.9  &  62.4  & 44.5   & 37.3   & 59.2   &   39.8 & 13.31 & 168.3 &  39.7   &  59.7  &  42.1  &  23.6  & \textbf{43.9} & 51.3\\
        & 35 & 23.82  & 186.7 & \textbf{41.1} &  \textbf{63.0}  & \textbf{44.8}   & \textbf{37.8}   & \textbf{59.8} & \textbf{40.3} & 13.32  & 168.4 & \textbf{40.1}  &  \textbf{60.5}  & \textbf{42.5} &  \textbf{24.4}  & 43.8 & \textbf{52.3} \\
        \hline
        \end{tabular}
    }
    \label{tab:object-detection-diff-kernel}
\end{table*}

\begin{table*}[t]
    \caption{Robustness on common corruption and out-of-distribution comparison among VAN-LKA-Tiny, and VAN-LSKA-Tiny with different kernel size on ImageNet-C and ImageNet-$\mathrm{\bar{C}}$. We report top-1 accuracy (IN-C), mean corruption error (mCE), and retention for ImageNet-C. We also report corruption error (IN-$\mathrm{\bar{C}}$) for ImageNet-$\mathrm{\bar{C}}$.}
    \centering
    \resizebox{2\columnwidth}{!}{
    \begin{tabular}{lcccc|ccc|cccc}
    \hline
    &&&&&&&&&&& \\
        \textbf{Model} &
        \textbf{Kernel size} &
        \textbf{Param (M)} &
        \textbf{GFLOPs} &
        \textbf{IN ($\uparrow$)} &
        \textbf{IN-C ($\uparrow$)} &
        \textbf{mCE ($\downarrow$)} &
        \textbf{Ret R ($\uparrow$)} &
        \textbf{IN-$\mathrm{\bar{\textbf{C}}}$ ($\downarrow$)} &
        \textbf{IN-A ($\uparrow$)} & 
        \textbf{IN-R ($\uparrow$)} &
        \textbf{IN-SK ($\uparrow$)}
        \\
        \hline
        \multirow{4}{*}{VAN-LKA-Tiny (Baseline)} & 7 & 4.02 & 0.84  & 74.0 & 44.1 & 71.5 & 59.6 & 53.6 & 5.3. & 38.0 & 25.9 \\
        & 11 & 4.04 & 0.85  & 74.4 & 45.6 & 69.5 & 61.3 & 51.5 & 6.5 & 38.6 & 25.9 \\
        & 23 & 4.11 & 0.87  &  74.7 & 46.0 & 69.0 & 61.5 & 50.5 & 7.1 & 39.1 & 26.2\\
        & 35 & 4.22 & 0.92  & \textbf{75.0}  & \textbf{46.6} & \textbf{68.2} & \textbf{62.2} & \textbf{49.8} & \textbf{7.8} & \textbf{39.3} & \textbf{26.4} \\
        \hline
        \multirow{4}{*}{VAN-LSKA-Tiny} & 7  & 4.01 & 0.83  & 73.5 & 44.0 & 71.7 & 59.9 & 53.7 & 4.7 & 37.8 & 25.3 \\
        & 11 & 4.02 & 0.84  & 74.5 & 45.4 & 69.9 & 60.9 & 51.5 & 6.3 & 38.6 & \textbf{26.2} \\
        & 23 & 4.03 & 0.84  & \textbf{75.0} & 46.3 & 68.6 & 61.9 & 50.6 & \textbf{7.5} & \textbf{38.5} & 25.7 \\
        & 35 & 4.04 & \textbf{0.85}  & 74.8 & \textbf{46.4} & \textbf{68.6} & \textbf{62.0} & \textbf{49.9} & 7.2 & 38.2 & 26.1  \\
        \hline
        \end{tabular}
        }
        \label{tab:robustness-comparsion-diff-kernel}
\end{table*}
\begin{table}[t]
    \caption{Semantic segmentation performance of VAN-LKA-Tiny and VAN-LSKA-Tiny with different kernel size on ADE20K validation set. Here "GFLOPs" is calculated with input size 2048x1024.}
    \centering
    \resizebox{1\columnwidth}{!}{
    \begin{tabular}{l|c|c|c|c}
        \hline
        \multirow{2}{*}{Backbone} & \multirow{2}{*}{Kernel size} & \multicolumn{3}{c}{Semantic FPN}\\ 
        \cline{3-5}
        & & Params(M) & GFLOPs & mIoU ($\%$) \\
        \hline
        \multirow{4}{*}{VAN-LKA-Tiny (Baseline)} & 7 & 7.89 & 205 & 38.3 \\
        & 11 & 7.92 & 205 & 38.5 \\
        & 23 & 7.98 & 206 & 40.0 \\
        & 35 & 8.10 & 208 & 39.9 \\
        & 53 & 8.36 & 213 & \textbf{40.5} \\
        \hline
        \multirow{4}{*}{VAN-LSKA-Tiny} & 7 & 7.89 & 205 & 38.4 \\
        & 11 & 7.89 & 205 & 38.8 \\
        & 23 & 7.91 & 205 & 39.5 \\
        & 35 & 7.92 & 205 & \textbf{40.2} \\
        & 53 & 7.94 & 206 & 39.3 \\
        \hline
        \end{tabular}
    }
    \label{tab:semantic-seg-diff-kernel}
\end{table}

We observed from the results of object detection, semantic segmentation, and robustness evaluation as shown in Table \ref{tab:object-detection-diff-kernel}, Table \ref{tab:semantic-seg-diff-kernel}, and Table \ref{tab:robustness-comparsion-diff-kernel} respectively, that the VAN-LKA-Tiny and the VAN-LSKA-Tiny obtained a higher accuracy when increasing the kernel size from $7\times7$ to $35\times35$.

For example, Table \ref{tab:object-detection-diff-kernel} shows increasing the kernel size of VAN-LSKA-Tiny from $7\times7$ to $35\times35$ improves mean average precision (mAP) by 1.6\% for Mask R-CNN. Similarly, we found that increasing the kernel size $7\times7$ to $35\times35$, in general, improves the semantic segmentation performance as shown in Table \ref{tab:semantic-seg-diff-kernel}. Similar trends are also observed for image classification under common corruptions (robustness evaluation) as shown in Table \ref{tab:robustness-comparsion-diff-kernel}.

\begin{table}[t]
    \caption{Percentage of neurons ($|z_k|/|z|$) encoding texture and shape (i.e., k) for the stage-4 latent representation. Note that the total dimension of the representation, $|z|$, is 256 and the remaining factors are assigned to the residual factor.}

    \centering
    \begin{tabular}{l|c|c|c}
        \hline
        \multirow{2}{*}{Backbone} & \multirow{2}{*}{Kernel size} & \multicolumn{2}{c}{Factor $|z_k|/|z|$ (\%)}\\ 
        \cline{3-4}
        & & Shape & Texture \\
        \hline
        \multirow{4}{*}{VAN-LKA-Tiny} & 7 & 18.9 & \textbf{31.3}\\
        & 11 & 19.0 & 31.0 \\
        & 23 & 19.4 & 30.4 \\
        & 35 & \textbf{19.5} & 30.1 \\
        \hline
        \multirow{4}{*}{VAN-LSKA-Tiny} & 7 & 18.7 & \textbf{31.8}\\
        & 11 & 18.9 & 31.4 \\
        & 23 & \textbf{19.8} & 29.5 \\
        & 35 & 19.7 & 29.9 \\
        \hline
        \end{tabular}
    \label{tab:shape-texture-ratio-w-diff-kernel}
\end{table}

\begin{table}[t]
    \caption{Texture classification performance (Top-1\%) of VAN-LKA-Tiny and VAN-LSKA-Tiny with different kernel sizes on DTD \cite{cimpoi2014describing} validation set.}

    \centering
    \begin{tabular}{l|ccccc}
        \hline
        Kernel size $\rightarrow$ & 7 & 11 & 23 & 35 & 53 \\
        \hline
        VAN-LKA-Tiny             & 51.8 & 51.3 & 51.6 & 50.5 & 41.7 \\
        VAN-LSKA-Tiny &  58.6 &58.7& 58.9 & 57.2 & 56.9 \\ 
        \hline
        \end{tabular}
    \label{tab:texture-analysis}
\end{table}

\subsection{Texture and shape bias with increasing kernel size}
\label{subsec:ab-studies-shape-texture}
Interestingly, we noted from Table \ref{tab:robustness-comparsion-diff-kernel} that the VAN-LKA-Tiny and VAN-LSKA-Tiny models with a larger kernel size provide better performance when introduced with corrupted data, which shows that VAN models with the larger kernel are more robust. A similar observation can be found in \cite{geirhos2018imagenet} and \cite{ding2022scaling}. We conjecture that such phenomena occur due to the VAN models becoming more biased to the shape than the texture of the object with increasing kernel size. It is also mentioned in \cite{geirhos2018imagenet} that the shape-based representation is more beneficial than texture-based representation in the downstream tasks since the ground truth labelings (i.e., bounding box and segmentation map) are aligned with the object shape. 

To verify that increasing the kernel size makes the VAN models more biased towards the shape than the texture, we follow the methodology in \cite{islam2021shape} to evaluate the amount of shape and texture information contained in the latent representation of the VAN-LKA-Tiny and VAN-LSKA-Tiny models. The main idea is that mutual information between two images, which share similar semantic properties (e.g., shape, texture, color, etc.), will be preserved in the latent representation only if the semantic properties are captured by the model. The calculation details can be found in \ref{shape-and-texture-dim-est}. In this experiment, we use the texture-shape cue conflict dataset \cite{geirhos2018imagenet} which consists of 1280 texture-shape cue conflict images generated by the style transfer \cite{gatys2016image} to form a pair of images that share similar semantic properties shape, or texture for estimating the dimensionality of two information. As shown in Table \ref{tab:shape-texture-ratio-w-diff-kernel}, VAN-LKA-Tiny and VAN-LSKA-Tiny encode more shape information as the kernel size increases while encoding lesser texture information. This result indicates that there is a correlation between the kernel size and the amount of shape information encoded. 

To investigate the impact of the LSKA and LKA's image texture analysis, we conducted a texture classification experiment using the Describable Textures Dataset (DTD) \cite{cimpoi2014describing} which consists of 5640 images with 47 categories. We performed texture classification tasks by employing the KNN classifier on the VAN-LKA-Tiny and VAN-LSKA-Tiny backbones with $K=1$, and the top-1 accuracy on the validation set was reported. As shown in Table \ref{tab:texture-analysis}, the performance of both LSKA and LKA decreased as the kernel size increased from 7 to 53. Notably, LSKA exhibited a comparatively minor performance drop than LKA when the kernel size increased from 35 to 53.

\section{Comparsion with state-of-the-art methods}
\label{sec:sota}
In this section, we compare VAN-LSKA with state-of-the-art CNN-based models (i.e., ConvNeXt \cite{liu2022convnet}, RepLKNet \cite{ding2022scaling}, SLaK \cite{liu2023more} and VAN-LKA \cite{guo2022visual}) and transformer-based models (i.e., DeiT \cite{touvron2021training}, PVT \cite{wang2021pyramid}, PVTv2 \cite{wang2022pvt}, and Swin Transformers \cite{liu2021swin}) on ImageNet \cite{deng2009imagenet} classification, ImageNet-C \cite{hendrycks2019benchmarking} robustness testing, COCO \cite{lin2014microsoft} object detection, and ADE20K \cite{zhou2017scene} semantic segmentation. To ensure a fair comparison with other ViTs and CNNs, we downscaled the RepLKNet-31B model from 79M parameters to the RepLKNet-31T model with 29.7M parameters. This downsizing involved reducing the channel size to 64, 128, 320, and 512 for model stages 1 to 4, respectively.
\begin{table*}[t]
    \caption{Comparisons with state-of-the-art methods on ImageNet-1K classification, COCO object detection and ADE20K semantic segmentation. The kernel size of VAN-LKA and VAN-LSKA is set to $k=35$ in the ImageNet classification task, COCO object detection, ADE20K semantic segmentation, and ImageNet-C robustness evaluation. We use the input size $224\times224$, $1280\times800$ and $512\times512$ for ImageNet-1K classification, COCO object detection and ADE20K semantic segmentation respectively when calculating the 'GFLOPs'.}
    \centering
    \resizebox{2\columnwidth}{!}{
    \begin{tabular}{l|ccc|c|ccc|ccc}
        \hline
        \multirow{2}{*}{Model} & 
        \multicolumn{3}{c|}{ImageNet} &
        IN-C &
        \multicolumn{3}{c|}{Mask R-CNN 1x COCO} &
        \multicolumn{3}{c}{Semantic FPN ADE20K}
        \\
         & Params.(M) &	GFLOPs & Top-1 & Top-1 & Params.(M) & GFLOPs & ${\rm AP}^b$ & Params.(M) & GFLOPs & mIoU ($\%$) \\
        \hline
        PVTv2-B0 \cite{wang2022pvt} & 3.4 & 0.6 & 70.5 & 40.2 & 23.5 & 178.6 & 38.2 & 7.6 & 24.0 & 37.2   \\
        DeiT-Tiny/16 \cite{touvron2021training} & 5.7  & 1.3 & 72.2 & 44.3 &  - & - & - & - & - & - \\
        VAN-LKA-Tiny \cite{guo2022visual} & 4.1  & 0.9 & 74.7 & \textbf{46.6} & 24.0 & 188.1 & \textbf{41.5} & 8.1 & 26.0 & 39.9 \\
        VAN-LSKA-Tiny (our) & 4.0 & 0.8 & \textbf{75.0} & 46.4 & 23.8 & 186.7 & 41.1 & 7.9 & 25.6 & \textbf{40.2}\\
        \hline
        PVT-Tiny \cite{wang2021pyramid} & 13.2  & 1.9 & 75.1 & 42.8 & 32.9 & 201.9 & 36.7 & 17.0 & 31.2 & 35.7 \\
        PVTv2-B1 \cite{wang2022pvt} & 13.1  & 2.1 & 78.7 & 50.1 & 33.7 & 205.7 & 41.8 & 17.8 & 32.1 & 41.5 \\
        VAN-LKA-Small \cite{guo2022visual} & 13.9 & 2.5 & 80.5 & 55.1 & 33.7 & 222.9 & 44.0 & 17.8 & 34.9 & 42.9 \\
        VAN-LSKA-Small (our) & 13.7 & 2.4 & \textbf{80.6} & \textbf{55.2} & 33.4 & 220.9 & \textbf{44.4} & 17.5 & 34.4 & \textbf{43.2}\\
        \hline
        DeiT-Small/16 \cite{touvron2021training} & 22.1 & 4.6 & 79.8 & 56.4 & - & - & - & - & - & - \\
        PVT-Small \cite{wang2021pyramid} & 24.5 & 3.8 & 79.8 & 52.3 & 44.1 & 232.2 & 40.4 & 28.2 & 40.6 & 39.8\\
        Swin-T \cite{liu2021swin} & 28.3 & 4.5 & 81.3 & 53.9 & 47.8 & 263.8 & 43.7 & 31.9 & 46.4 & - \\
        PVTv2-B2 \cite{wang2022pvt} & 25.4 & 4.0 & 82.0 & 58.1 & 45.0 & 237.0 & 45.3 & 29.1 & 41.9 & 45.2 \\
        ConvNeXt-T \cite{liu2022convnet} & 28.6 & 4.5 & 82.1 & 58.0 & 48.1 & 262.1 & - & 32.2 & 45.0 & -\\
        SLaK-T \cite{liu2023more} & 30.0 & 5.0 & 82.5 & 58.9 & 50.2 & 278.8 & 44.8 & 34.5 & 49.9 & 44.4 \\
        RepLKNet-31T \cite{ding2022scaling} & 29.7 & 5.9 & 82.5 & 58.1 & 49.3 & 292.5 & 39.6 & 33.4 & 52.7 & 43.9 \\
        VAN-LKA-Base \cite{guo2022visual} & 26.6 & 5.0 & 82.8 & \textbf{60.7} & 46.7 & 275.5 & \textbf{46.1} & 30.8 & 48.4 & 45.6\\
        VAN-LSKA-Base (our) & 26.2 & 4.9 & \textbf{82.8} & 60.3 & 46.0 & 271.5 & 45.9 & 30.1 & 47.4 & \textbf{46.9}\\
        \hline
        \end{tabular}
    }
    \label{tab:sota-classification}
\end{table*}

\begin{table}[t]
    \caption{Percentage of neurons ($|z_k|/|z|$) encoding texture and shape (i.e., k) for the stage-4 latent representation with ViTs and CNNs}

    \centering
    \begin{tabular}{l|c|c}
        \hline
        \multirow{2}{*}{Backbone} & \multicolumn{2}{c}{Factor $|z_k|/|z|$ (\%)}\\ 
        \cline{2-3}
        & Shape & Texture \\
        \hline
        DeiT-Small/16 \cite{touvron2021training} & 19.1 & 31.0\\
        PVT-Small \cite{wang2021pyramid} & 18.6 & 31.2 \\
        Swin-T \cite{liu2021swin} & 18.4 & \textbf{32.3}\\
        PVTv2-B2 \cite{wang2022pvt} & 19.1 & 29.9\\
        ConvNeXt-T \cite{liu2022convnet} & 19.4 & 28.9 \\
        SLaK-T \cite{liu2023more} & 19.3 & 29.0 \\
        RepLKNet-31T \cite{ding2022scaling} & 19.0 & 29.9 \\
        VAN-LKA-Base (k=35) \cite{guo2022visual} & 19.5 & 29.4 \\
        VAN-LSKA-Base (k=35) & \textbf{19.9} & 28.8 \\
        \hline
        \end{tabular}
    \label{tab:shape-texture-ratio-w-diff-model}
\end{table}

\begin{table*}[h]

    \caption{Robustness and out-of-distribution comparison with SOTA methods on ImageNet-C, ImageNet-A, ImageNet-R, and ImageNet-Sketch.}
    \centering
    \resizebox{2\columnwidth}{!}{
    \begin{tabular}{l|l|ccc|ccc|cccc}
        \hline
        \textbf{Type} &
        \textbf{Model} &
        \textbf{Param (M)} &
        \textbf{GFLOPs} &
        \textbf{IN ($\uparrow$)} &
        \textbf{IN-C ($\uparrow$)} &
        \textbf{mCE ($\downarrow$)} &
        \textbf{Ret R ($\uparrow$)} &
        \textbf{IN-$\mathrm{\bar{\textbf{C}}}$ ($\downarrow$)} &
        \textbf{IN-A ($\uparrow$)} &
        \textbf{IN-R ($\uparrow$)} &
        \textbf{IN-SK ($\uparrow$)} \\
        \hline
        \multirow{4}{*}{ViTs} & DeiT-S/16 \cite{touvron2021training} & 22.1 & 4.6 & 79.8 & 56.4 & 55.8 & 70.6 & 39.4 & 19.8 & 41.9 & 29.1 \\
        & PVT-Small \cite{wang2021pyramid} & 24.5 & 3.8 & 79.8 & 52.3 & 61.1 & 65.5 & 44.4 & 18.0 & 39.7 & 27.0\\
        & Swin-T \cite{liu2021swin} & 28.3 & 4.5 & 81.2 & 53.9 & 59.0 & 66.3 & 42.7 & 21.6 & 41.3 & 29.0\\
        & PVTv2-B2 \cite{wang2018non}  & 25.4 & 4.0  &  82.0 & 58.1 & 53.8 & 70.8 & 38.5 & 27.9 & 45.1 & 32.7\\
        \hline
        \multirow{5}{*}{CNN} & ConvNeXt-T \cite{liu2022convnet} & 28.6 & 4.5 & 82.1 & 58.0 & 53.8 & 70.6 & 41.3 & 24.2 & 47.1 & 33.9 \\
        & SLaK-T \cite{liu2023more} & 30.0 & 5.0 & 82.5 & 58.9 & 52.7 & 71.4 & 40.1 & 29.9 & 45.3 & 32.4 \\
        & RepLKNet-31T \cite{ding2022scaling} & 29.7 & 5.9 & 82.5 & 58.1 & 53.6 & 70.4 & 40.5 & 30.3 & 47.5 & 35.0 \\
        & VAN-LKA-Base (k=35) \cite{guo2022visual} & 26.6 & 5.0 & 82.8 & \textbf{60.7} & \textbf{50.3} & \textbf{73.3} & 37.5 & \textbf{31.8} & \textbf{47.6} & \textbf{35.6}\\
        & VAN-LSKA-Base (k=35) (Ours)  & 26.2 & 4.9 & \textbf{82.8} & 60.3 & 51.0 & 72.8 & \textbf{36.6} & 30.1 & 46.6 & 34.6\\
        
        \hline
        \end{tabular}
        }
    \label{tab:robustness-comparsion-diff-model}
\end{table*}

As shown in Table \ref{tab:sota-classification}, the Tiny, Small, and Base versions of our proposed VAN-LSKA network outperformed ViTs in the category of ImagNet classification, object detection, semantic segmentation, and robustness evaluation. For instance, the proposed method surpasses Swin Transformer (Swin-T) by 1.5\% accuracy on classification while saving 7.4\% parameters. This performance gap further enlarges by 6.4\% and 2.2\% on robustness evaluation, and object detection respectively. Our proposed method also surpasses large kernel-based ConvNext by 0.7\% and 2.3\% on ImageNet classification and ImageNet-C robustness evaluation respectively while saving 8.4\% parameters. Compared to the recent state-of-the-art methods, RepLKNet and SLaK, our method outperform both methods by 0.3\% on ImageNet classification in accuracy. Furthermore, on the ImageNet-C dataset, our method achieves improvements of 2.2\% and 1.4\% over RepLKNet and SLaK, respectively. Notably, our method can save 11.7\% and 12.6\% parameter sizes compared to both methods. We conjecture that the performance difference between the proposed VAN-LSKA method and state-of-the-art ViTs and CNN architectures is due to LSKA module in VAN enabling the large receptive field (e.g., $35\times 35$). Such use of the large kernel size improves the shape bias while lowering the texture bias which has been shown to improve the performance as mentioned in subsection \ref{subsec:ab-studies-shape-texture} and also in \cite{ding2022scaling, geirhos2018imagenet}. We report the texture and shape dimensionality for VAN-LSKA, VAN-LKA, ViTs, and CNN-based architectures in Table \ref{tab:shape-texture-ratio-w-diff-model}. One can see from Table \ref{tab:shape-texture-ratio-w-diff-model}, that VAN-LSKA (with kernel size $35\times 35$) has a higher shape and lower texture dimensionality than VAN-LKA, ConvNeXt, RepLKNet, SLaK and ViTs supporting the evidence of better performance with large-kernel size. 

When compared to VAN-LKA, the proposed method achieves comparable performance on classification, robustness evaluation, object detection, and semantic segmentation tasks. However, Compared to VAN-LKA, VAN-LSKA obtains a lower parameter size and GFLOPs.

\section{Benchmarking Robustness Evaluation on ViTs and CNNs}
\label{sec:sota-robust}
We further provide the robustness comparison among ViTs and CNNs as shown in Table \ref{tab:robustness-comparsion-diff-model}. We follow the evaluation steps as mentioned in session \ref{sec:robustness-eval} and directly test the pre-trained ImageNet1k models on ImageNet-$\mathrm{\bar{C}}$ \cite{mintun2021interaction}, ImageNet-A \cite{hendrycks2021natural}, ImageNet-R \cite{hendrycks2021many}, and ImageNet-SK \cite{wang2019learning}. Compared to ViTs, VAN-LSKA outperforms all the state-of-the-art transformer-based models in all datasets. This result inlines with our findings in the previous session \ref{sec:sota}. Compared to CNNs, VAN-LSKA achieves better performance than ConvNeXt, RepLKNet, and SLaK in most of the datasets while saving 8.4\%, 11.7\% and 12.6\% parameters, respectively. When compared to the VAN-LKA model, our model achieves comparable performance with lower parameter sizes and GFLOPs.

\section{Conclusion}
\label{sec:conclusion}
In this paper, we addressed the issue of computational inefficiency of the depth-wise convolution when increasing the kernel size in the LKA modules. To mitigate these issues, we proposed a simple strategy with the use of cascaded horizontal and vertical one-dimensional depth-wise convolution which effectively reduces the quadratic growth in the number of parameters and computational flops incurred by the depth-wise convolution in the LKA-trivial and LKA in VAN. Experimental results demonstrate that the proposed LSKA in VAN can achieve a good trade-off between kernel size, parameter size, and speed while maintaining a comparable or better performance than LKA-trivial and LKA in various computer vision tasks. We further study the robustness of LSKA and demonstrate that LSKA is a robust learner compared to previous large kernel CNNs and ViTs. We provide quantitative evidence to show that increasing the kernel size of LSKA enables the model to encode more shape information and less texture in the feature representation. Since there is a high correlation between the quantity of shape information encoded in the representation and robustness, this helps us understand why LSKA-based VAN is a robust learner. On the other hand, our work may open a new direction for exploring the benefit of LSKA in other domains like video tasks or acoustic tasks in the future.

\section*{Acknowledgments}
We would like to thank Yujia Zhang for running part of the experiments. We thank Yuzhi Zhao and Wing-Yin Yu for their kind suggestions.

\appendix[]
\section*{Shape and Texture Dimensionality Estimation Method}
\label{shape-and-texture-dim-est}
\begin{figure}[h]
\includegraphics[width=\linewidth]{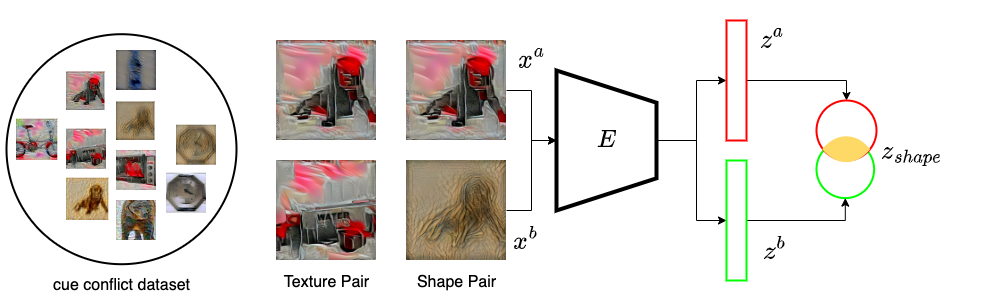}
\caption{Illustration of the shape and texture dimensionality estimation procedure. We first sample image pairs that share the same shape (right pair) and texture (left pair) from the cue conflict dataset \cite{geirhos2018imagenet}. We then feed these pairs into the ImageNet1k pre-trained model $E$ and estimate the mutual information of the representation, $z^a$ and $z^b$, corresponding to shape $z_{shape}$ and texture $z_{texture}$. }
\label{fig:shape-texture-measurement}
\end{figure}
We follow the methodology in \cite{islam2021shape} to estimate the shape and texture information encoded in the latent representation. Given an image pair $(x^a,x^b)$ that share a similar semantic factor (i.e., shape or texture), our aim is to estimate the dimensionality of the semantic factor in the latent vector $z$ which is encoded by a pre-trained model $E$. The mutual information of the image pair, which is shape or texture, will be preserved in the representation $z$ only if the model $E$ can capture them. Thus, based on the simple assumption that the joint distribution for neurons $z^a_{i}$ and $z^b_{i}$ in latent representations are bivariate normal distribution, we can estimate their mutual information (MI) with their correlation coefficient $c_{i}$ as follow:
\begin{equation}
\begin{split}
   MI(z^a_{i}, z^b_{i})=-\dfrac{1}{2}log(1-c_{i}^2), \\ 
\text{where}~~ c_{i}=\dfrac{Cov(z^a_{i},z^b_{i})}{\sqrt{Var(z^a_{i})Var(z^b_{i})}} 
\end{split}
\end{equation}
To estimate the number of neurons that represent shape and texture, we sum all of their respective correlation coefficients $\sum_ic^{shape}_{i}$ and $\sum_ic^{texture}_{i}$ as final score $s_{shape}$ and $s_{texture}$. Since the correlation is bounded by $[-1, 1]$, the score $s_{shape}$ and $s_{texture}$ are in $[-N, N]$, where $N$ is the dimension of the latent representation. In order to ensure that the sum of all dimensions of the semantic factors is equal to the dimension of the latent representation $N$, we apply a softmax function on the final score vector $s_{k}$, where $k$ represents the semantic factors (i.e., shape and texture). Note that the remaining dimensions that are not included in these two semantic factors are allocated as residual semantic factors. Finally, the dimensionality of the semantic factors $N_{k}$ can be calculated as follow:
\begin{equation}
    N_{k}=\dfrac{e^{s_{k}}}{\sum^K_{j=0}e^{s_{j}}} \times N
\end{equation}
where $K=3$ represents the shape, texture, and residual factors. Fig. \ref{fig:shape-texture-measurement} shows the detail of the shape and texture dimensionality estimation procedure.

\section*{Effective Receptive Field Comparison Between LKA and LSKA beyond kernel size of 23}
\label{erf-comp}
\begin{figure*}[t]
      \centering
      \subfloat[VAN-LSKA-Tiny (k=35)]{\includegraphics[width=0.25\linewidth, height=0.25\linewidth]{van_lsk_w_dilation_tiny_k35_erf.png}
      \label{fig:erf_lska_k35}}
      \subfloat[VAN-LSKA-Tiny (k=53)]{\includegraphics[width=0.25\linewidth,height=0.25\linewidth]{van_lsk_w_dilation_tiny_k53_erf.png}
        \label{fig:erf_lska_k53}}
      \subfloat[VAN-LSKA-Tiny (k=65)]{\includegraphics[width=0.25\linewidth,height=0.25\linewidth]{van_lsk_w_dilation_tiny_k65_erf.png}
        \label{fig:erf_lska_k65}}\\
      \subfloat[VAN-LKA-Tiny (k=35)]{\includegraphics[width=0.25\linewidth,height=0.25\linewidth]{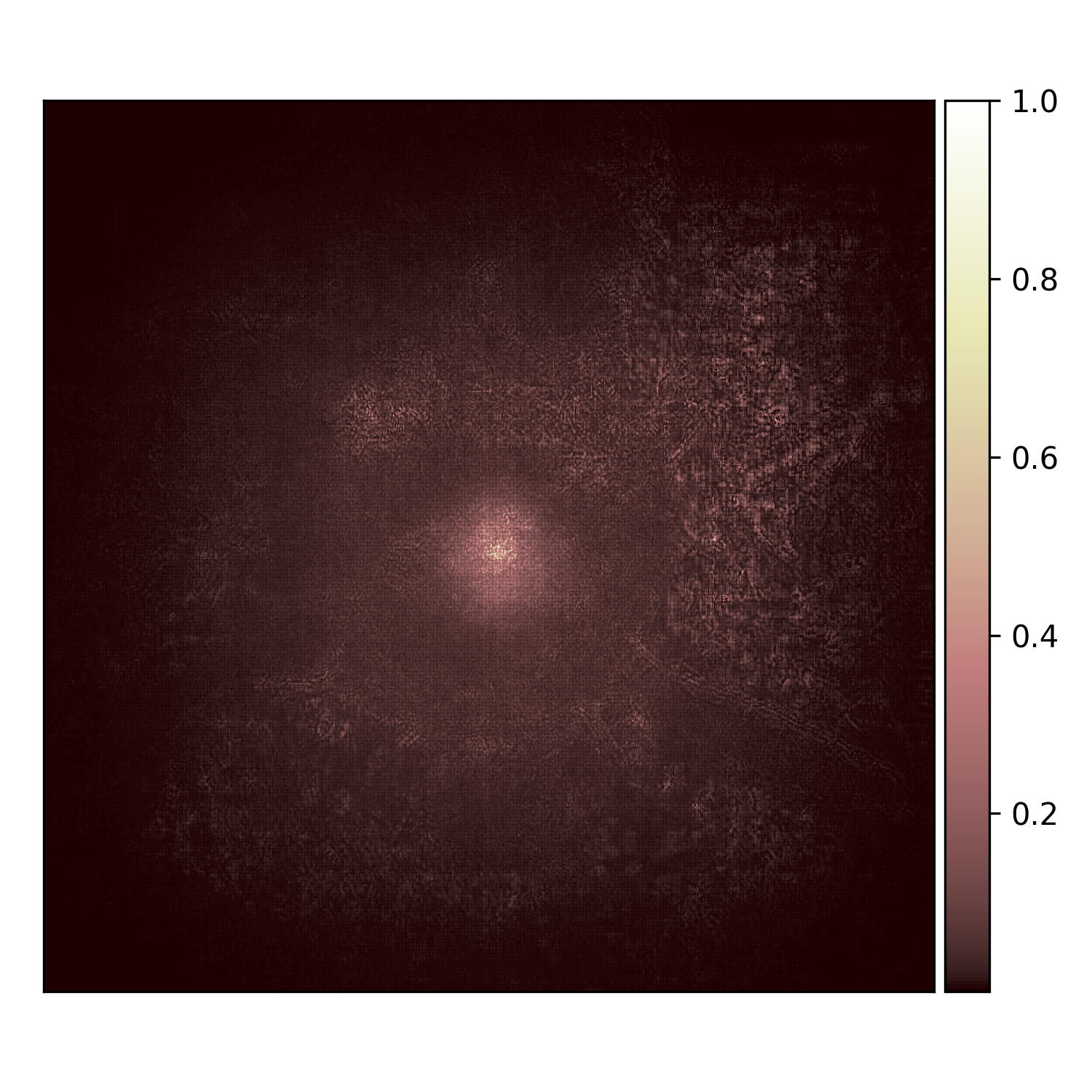}
          \label{fig:erf_lka_k35}
        }
        \subfloat[VAN-LKA-Tiny (k=53)]{\includegraphics[width=0.25\linewidth,height=0.25\linewidth]{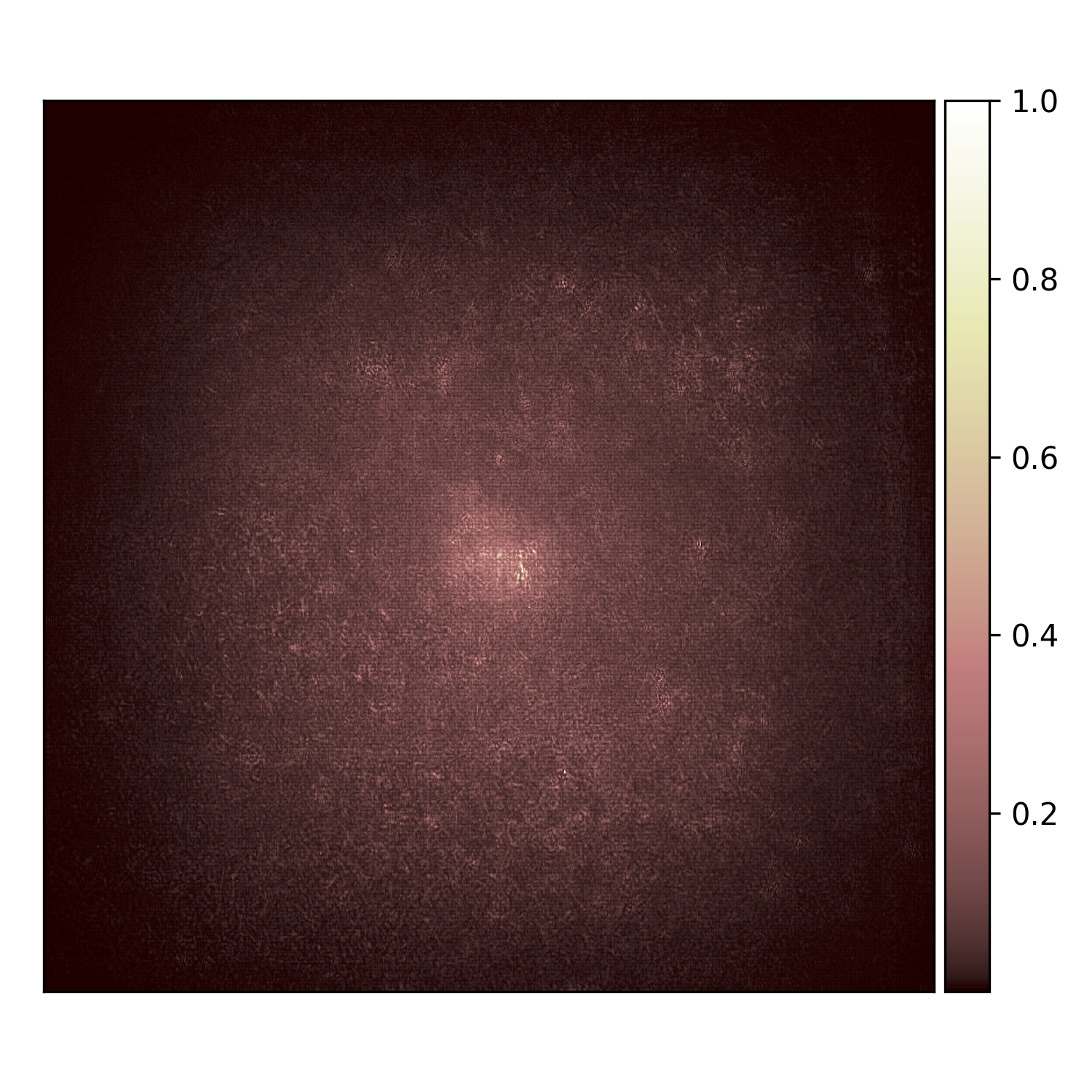}
              \label{fig:erf_lka_k53}
            }
        \subfloat[VAN-LKA-Tiny (k=65)]{\includegraphics[width=0.25\linewidth,height=0.25\linewidth]{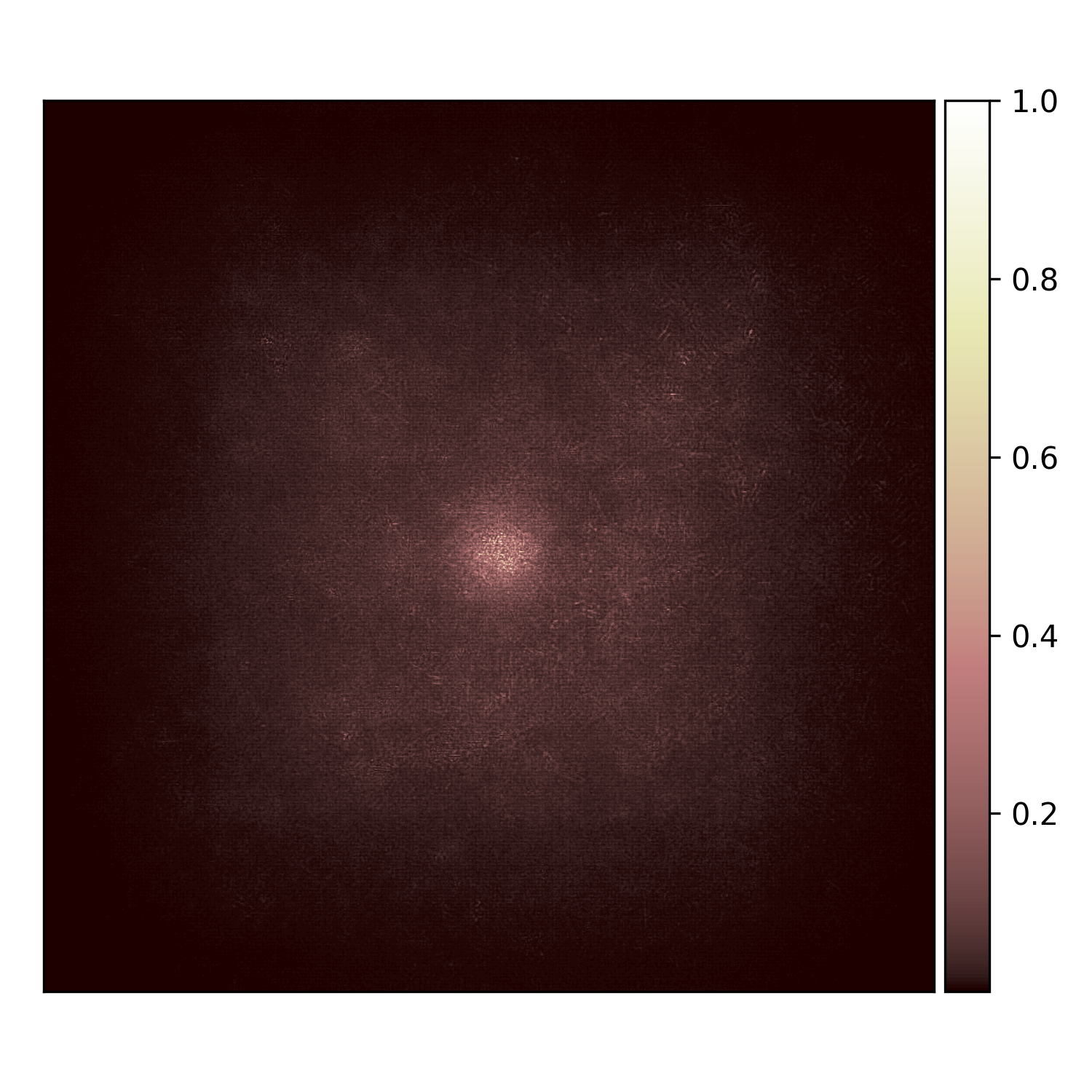}
              \label{fig:erf_lka_k65}
            }
      \caption{Effective Receptive Field (ERF) of VAN-LSKA-Tiny and VAN-LKA-Tiny with kernel size 35, 53, and 65. The ERF of LSKA reaches saturation beyond the kernel sizes of 35, 53, and 65, similar to LKA.}
      \label{fig:erf-lska-lka}
\end{figure*}

See Fig.\ref{fig:erf-lska-lka}.

\section*{Comaprsion among LKA-trivial, LKA, LSKA-trivial, and LSKA on Tiny models with different kernel size}
See Table \ref{tab:performance-comp-tiny}.

\begin{table*}[t]
    \caption{Comaprsion among LKA-trivial, LKA, LSKA-trivial, and LSKA with different kernel sizes on ImageNet classification. Note that we use Tiny models to conduct the comparison.}
    \centering
    \resizebox{2\columnwidth}{!}{
    \begin{tabular}{l|cccc|cccc|cccc|cccc}
        \hline
         Backbone & \multicolumn{4}{c|}{LKA-trival} & \multicolumn{4}{c|}{LSKA-trival} & \multicolumn{4}{c|}{LKA} & \multicolumn{4}{c}{LSKA}\\ 
         \hline
         Kernel Size & Params(M) & GFLOPs & Inf. Speed & Top-1 & Params(M) & GFLOPs & Inf. Speed & Top-1 & Params(M) & GFLOPs & Inf. Speed & Top-1 & Params(M) & GFLOPs & Inf. Speed & Top-1 \\
        \hline
        7-7-7-7 & 4.06 & 0.86 & 711 & 74.2 & 4.01 & 0.83 & 733 & 73.8 & 4.02 & 0.84 & 745 & 74.0  & 4.01 & 0.83 & 756 & 73.5 \\
        11-11-11-11 & 4.18 & 0.90 & 630 & 74.7 & 4.03 & 0.84 & 719 & 74.8 & 4.04 & 0.85 & 709 & 74.4 & 4.02 & 0.84 & 746 & 74.5 \\
        23-23-23-23 & 4.83 & 1.16 & 374 & 75.2 & 4.06 & 0.85 & 678 & 75.0 & 4.11 & 0.87 & 660 & 74.7 & 4.03 & 0.84 & 713 & 75.0  \\
        35-35-35-35 & 5.95 & 1.60 & 243 & 75.1 & 4.10 & 0.87 & 677 & 74.8 & 4.22 & 0.92 & 594 & 75.0 & 4.04 & 0.85 & 684 & 74.8 \\
        53-53-53-53 & 8.15 & 2.61 & 146 & 75.1 & 4.16 & 0.89 & 599 & 74.9 & 4.49 & 1.02 & 511 & 75.1 & 4.06 & 0.85 & 656 & 74.8 \\
        65-65-65-65 & 10.74 & 3.49 & 112 & 75.1 & 4.19 & 0.90 & 556 & 75.0 & 4.73 & 1.12 & 451 & 75.1 & 4.07 & 0.85 & 650 & 74.8 \\
        \hline 
        \end{tabular}
    }
    \label{tab:performance-comp-tiny}
\end{table*}

\bibliographystyle{IEEEtran}
\bibliography{egbib}

\vfill

\end{document}